\documentclass{article}

\PassOptionsToPackage{numbers, compress}{natbib}
\usepackage{natbib,fullpage}

\usepackage[colorlinks=true,
linkcolor=red,
urlcolor=blue,
citecolor=blue]{hyperref}

\usepackage[utf8]{inputenc} 
\usepackage[T1]{fontenc}    
\usepackage{hyperref}       
\usepackage{url}            
\usepackage{booktabs}       
\usepackage{amsfonts}       
\usepackage{nicefrac}       
\usepackage{microtype}      
\usepackage{shortcuts_OPT}
\usepackage{bm,amsmath,multicol,caption}
\usepackage{algorithm,algorithmic}
\usepackage{xargs}
\usepackage{cleveref}
\usepackage{graphicx}
\usepackage{longtable}
\usepackage{answers}
\usepackage{array}
\usepackage{amsthm}

\newtheorem{Lemma}{Lemma}

\newtheorem{Theorem}{Theorem}

\newtheorem{assumption}{A\!\!}

\newtheorem*{Theorem*}{Theorem}

\newcommand{\xupdownarrow}[1]{%
  {\left\updownarrow\vbox to #1{}\right.\kern-\nulldelimiterspace}
}

\usepackage[textsize=footnotesize]{todonotes} 

\title{Multi-Agent Reinforcement Learning via Double Averaging Primal-Dual Optimization}

\author{
  Hoi-To Wai, Zhuoran Yang, Zhaoran Wang, Mingyi Hong\thanks{H.-T. Wai is with Chinese University of Hong Kong, Hong Kong. 
  Z. Yang is with Princeton University, NJ, USA. Z. Wang is with Northwestern University, IL, USA. M. Hong is with University of Minnesota, MN, USA. E-mails: \texttt{htwai@se.cuhk.edu.hk}, \texttt{zy6@princeton.edu}, \texttt{zhaoranwang@gmail.com}, \texttt{mhong@umn.edu}}
}

\begin{document}

\maketitle

\begin{abstract}
Despite the success of single-agent reinforcement learning, multi-agent reinforcement learning (MARL) remains challenging due to complex interactions between agents. Motivated by decentralized applications such as sensor networks, swarm robotics, and power grids, we study policy evaluation in MARL, where agents with jointly observed state-action pairs and private local rewards collaborate to learn the value of a given policy.
In this paper, we propose a double averaging scheme, where each agent iteratively performs averaging over both space and time to incorporate neighboring gradient information and local reward information, respectively. We prove that the proposed algorithm converges to the optimal solution at a global geometric rate. In particular, such an algorithm is built upon a primal-dual reformulation of the mean squared projected Bellman error minimization problem, which gives rise to a decentralized convex-concave saddle-point problem. To the best of our knowledge, the proposed double averaging primal-dual optimization algorithm is the first to achieve fast finite-time convergence on decentralized convex-concave saddle-point problems. 
\end{abstract}

\section{Introduction}
Reinforcement learning combined with deep neural networks recently achieves superhuman performance on various challenging tasks such as video games and board games \citep{mnih2015human, silver2017mastering}. In these tasks, an agent uses deep neural networks to learn from the environment and adaptively makes optimal decisions. Despite the success of single-agent reinforcement learning, multi-agent reinforcement learning (MARL) remains challenging, since each agent interacts with  not only the environment but also   other agents. 
In this paper, we study collaborative MARL with local rewards. In this setting, all the agents share a joint state whose transition dynamics is determined together by the local actions of individual agents. However, each agent only observes its own reward, which may differ from that of other agents. The agents aim to collectively maximize the global sum of local rewards. 
To collaboratively make globally optimal decisions, the agents need to exchange local information. Such a setting of MARL is ubiquitous in large-scale applications such as sensor networks \citep{rabbat2004distributed, cortes2004coverage}, swarm robotics \citep{kober2012reinforcement, corke2005networked}, and power grids \citep{callaway2011achieving, dall2013distributed}. \\
A straightforward idea is to set up a central node that  collects and broadcasts the reward information, and assigns the action of each agent.  This reduces the multi-agent problem into a single-agent one. However, the central node is often unscalable, susceptible to malicious attacks,   and even infeasible in large-scale applications. Moreover, such a central node is a single point of failure, which is susceptible to adversarial attacks. In addition, the agents are likely to be reluctant to reveal their local reward information due to privacy concerns \citep{chaudhuri2011differentially,lin14privacy}, which makes the central node unattainable.\\
To make MARL more scalable and robust, we propose a decentralized scheme for exchanging local information, where each agent only communicates with its neighbors over a network. In particular, we study the policy evaluation problem, which aims to learn a global value function of a given policy. We focus on minimizing a Fenchel duality-based reformulation of the mean squared Bellman error in the model-free setting with infinite horizon, batch trajectory, and linear function approximation.\\
At the core of the proposed algorithm is a ``double averaging" update scheme, in which the algorithm performs one average over space (across agents to ensure consensus) and  
one over time (across observations along the trajectory). In detail, each agent locally tracks an estimate of the full gradient and incrementally updates it using two sources of information: (i) the stochastic gradient evaluated on a new pair of joint state and action along the trajectory and the corresponding local reward, and (ii) the local estimates of the full gradient tracked by its neighbors. Based on the updated estimate of the full gradient, each agent then updates its local copy of the primal parameter. By iteratively propagating the local information through the network, the agents reach global consensus and collectively attain the desired primal parameter, which gives an optimal approximation of the global value function. 

\textbf{Related Work}~~The study of MARL in the context of Markov game dates back to \cite{littman1994markov}. See also \cite{littman2001value, lauer2000algorithm, hu2003nash} and recent works on collaborative MARL \cite{wang2003reinforcement,arslan2017decentralized}. However, most of these works consider the tabular setting, which suffers from the curse of dimensionality. 
To address this issue,   under the collaborative  MARL framework,   
 \cite{zhang2018fully}
  and 
\cite{lee2018primal} study actor-critic algorithms and  policy evaluation with on linear function approximation, respectively. However, 
their analysis is asymptotic in nature and largely relies on two-time-scale stochastic approximation using ordinary differential equations \citep{borkar2008stochastic}, which is tailored towards the continuous-time setting. Meanwhile, most works on collaborative MARL impose the simplifying assumption that the local rewards are identical across agents, making it unnecessary to exchange the local information. 
More recently, \cite{foerster2016learning,foerster2017stabilising, gupta2017cooperative,lowe2017multi,omidshafiei2017deep} study deep MARL that uses deep neural networks as function approximators. However, most of these works focus on empirical performance and lack theoretical guarantees. Also, they do not emphasize on the efficient exchange of information across agents. In addition to MARL, another line of related works study multi-task reinforcement learning (MTRL), in which an agent aims to solve multiple reinforcement learning problems with shared structures \citep{wilson2007multi,parisotto2015actor, macua2015distributed, macua2017diff, teh2017distral}.\\
The primal-dual formulation of reinforcement learning is studied in \cite{liu2015finite, macua2015distributed, macua2017diff, lian2016finite, dai2016learning,chen2016stochastic, wang2017primal, dai2017smoothed, dai2017boosting, du2017stochastic} among others. Except for \cite{macua2015distributed, macua2017diff} discussed above, most of these works study the single-agent setting. 
Among them, \cite{lian2016finite, du2017stochastic} are most related to our work. In specific, they develop variance reduction-based algorithms \citep{johnson2013accelerating, defazio2014saga, schmidt2017minimizing} to achieve the geometric rate of convergence in the setting with batch trajectory. In comparison, our algorithm is based on the aforementioned double averaging update scheme, which updates the local estimates of the full gradient using both the estimates of neighbors and new states, actions, and rewards. In the single-agent setting, our algorithm is closely related to stochastic average gradient (SAG) \citep{schmidt2017minimizing} and stochastic incremental gradient (SAGA) \citep{defazio2014saga}, with the difference that our objective function is a finite sum  
convex-concave saddle-point problem.\\
Our work is also related to prior work in the broader contexts of 
primal-dual and multi-agent optimization. For example, \cite{palaniappan2016stochastic} apply variance reduction techniques to convex-concave saddle-point problems to achieve the geometric rate of convergence. However, their algorithm is centralized and it is unclear whether their approach is readily applicable to the multi-agent setting. Another line of related works study multi-agent optimization, for example, \cite{tsitsiklis1986distributed,nedic2009distributed,chen2012diffusion,shi2015extra,qu2017harnessing}.  However, these works mainly focus on the general setting where the objective function is a sum of convex local cost functions. To the best of our knowledge, our work is the first to address decentralized convex-concave saddle-point problems with sampled observations that arise from MARL.


\textbf{Contribution}~~Our contribution is threefold: (i) We reformulate the multi-agent policy evaluation problem using Fenchel duality and propose a decentralized primal-dual optimization algorithm with a double averaging update scheme. (ii) We establish the global geometric rate of convergence for the proposed algorithm, making it the first algorithm to achieve fast linear convergence for MARL. (iii) Our proposed algorithm and analysis is of independent interest for solving a broader class of decentralized  convex-concave saddle-point problems with sampled observations.

\textbf{Organization}~~In \S\ref{sec:pf} we introduce the problem formulation of MARL. In \S\ref{sec:iag} we present the proposed algorithm and lay out the convergence analysis. In \S\ref{sec:ne} we illustrate the empirical performance of the proposed algorithm. We defer the detailed proofs to the supplementary material.

\textbf{Notation}~~Unless otherwise specified, for a vector ${\bm x}$, $\| {\bm x} \|$ denotes its Euclidean norm; for a matrix ${\bm X}$, $\| {\bm X} \|$ denotes its spectral norm, \ie the largest singular value.\vspace{-.4cm}


\section{Problem Formulation}\label{sec:pf}\vspace{-.2cm}
In this section, we  introduce the background of MARL, which is modeled as a multi-agent Markov decision process (MDP). Under this model, we formulate the policy evaluation problem as a primal-dual convex-concave optimization problem.

\textbf{Multi-agent MDP}~~Consider a group of $N$ agents. 
We are interested in the multi-agent MDP: 
\begin{align*}
\big(  {\cal S} , \{ {\cal A}_i \}_{i = 1}^N, {\cal P}^{\bm a}, \{ {\cal R}_i \}_{i=1}^N, \gamma \big) \eqs,
\end{align*}
where ${\cal S} $ is the state space and ${\cal A}_i$
is the action space for agent $i$. We write 
${\bm s} \in {\cal S}$
and ${\bm a} \eqdef (a_1,...,a_N) \in {\cal A}_1 \times \cdots \times {\cal A}_N$ 
as the joint state and action, respectively. 
The function ${\cal R}_i ({\bm s}, {\bm a})$ is the 
local reward received by agent $i$
after taking joint action
${\bm a}$ at state ${\bm s} $, and  $\gamma \in (0,1)$ is the discount factor.  
Both ${\bm s}$ and ${\bm a}$ are available to all agents, whereas the reward ${\cal R}_i$ is  {\it private} for agent $i$.

In contrast to a single-agent MDP, the agents are coupled
together by the state transition matrix 
${\cal P}^{\bm a} \in \RR^{   |{\cal S}| 
\times  |{\cal S}| }$, whose $({\bm s}, {\bm s}')$-th element 
is the probability of transiting from ${\bm s}$ to ${\bm s}'$,  
 after taking a  joint action  ${\bm a}$.
This scenario arises from large-scale applications such as sensor networks \citep{rabbat2004distributed, cortes2004coverage}, swarm robotics \citep{kober2012reinforcement, corke2005networked}, and power grids \citep{callaway2011achieving, dall2013distributed}, which strongly motivates the development of a multi-agent RL strategy. 
Moreover,  under the collaborative setting, the goal is to maximize the collective return of all agents.  Suppose there exists a central controller that collects the rewards of, and assigns the action to  each individual agent, the problem reduces to the classical MDP with action space $\cal A$ and global reward function $R_c ({\bm s}, {\bm a} ) = N^{-1} \sum_{i=1}^N {\cal R}_i ({\bm s}, {\bm a})$.  Thus, without such a central controller, it is essential for the agents to collaborate with each other so as to solve the multi-agent problem based solely on  local information.

Furthermore, a joint policy, denoted by $\bm{\pi}$, specifies the rule of making sequential decisions for the agents.  Specifically, $\bm{\pi} ( {\bm a} | {\bm s}  )$ is the  conditional probability of taking joint action ${\bm a} $ given the 
current state ${\bm s} $. 
We define the reward function of joint policy $\bm{\pi}  $  as an average of the local rewards:
\beq \label{eq:reward} \textstyle
R_c^{\bm{\pi}} ( {\bm s} ) \eqdef \frac{1}{N} \sum_{i=1}^N R_i^{\bm \pi} ({\bm s} ), \qquad \text{where}~~ 
R_i^{{\bm \pi }} ({\bm s} )  \eqdef 
\EE_{ {\bm a} \sim {\bm \pi} ( \cdot | {\bm s}  ) } \big[  {\cal R}_i ( {\bm s}, {\bm a} ) \big] \eqs.
\eeq
That is, $R_c^{\bm {\pi}}({\bm s} )  $ is the expected value of the average of the rewards when the agents follow policy ${\bm \pi}$ at state ${\bm s}$.
Besides,  any fixed policy ${\bm \pi}$ induces a Markov chain over $\cal S$, whose transition matrix is  denoted by ${\bm P} ^{\bm \pi}$.  The $({\bm s}, {\bm s}')$-th element  of ${\bm P} ^{\bm \pi}$ is given by 
\begin{align*} \textstyle
[ {\bm P} ^{\bm \pi}]_{{\bm s}, {\bm s}' }  = \sum_{{\bm a} \in {\cal A} }  {\bm \pi }({\bm a}  | {\bm s} ) \cdot [ {\cal P}^{\bm a} ]_{{\bm s}, {\bm s}' } .
\end{align*} 
When this Markov chain is aperiodic and irreducible, it induces a stationary distribution $\mu^{\bm \pi}$  over $\cal S$.

\textbf{Policy Evaluation}~~
A central problem in reinforcement learning is \emph{policy evaluation}, which refers to learning the \emph{value function} of a given policy. 
This problem appears as a key component in both value-based methods such as policy iteration, and policy-based methods such as actor-critic algorithms \citep{sutton1998reinforcement}. 
Thus, efficient estimation of  the value functions in  multi-agent MDPs enables us to extend the successful approaches in single-agent RL to the setting of MARL.

Specifically, 
for any given joint policy $\bm{\pi}$, the value function of ${\bm \pi} $, denoted by $V^{\bm{\pi}} \colon \cal S\rightarrow \RR$, is defined as the expected value of the discounted cumulative reward when the  multi-agent MDP is  initialized with a given state and the agents follows policy ${\bm \pi}$ afterwards. For any state
 ${\bm s} \in \cal S$, we define
\beq \label{eq:loc_reward} \textstyle
V^{\bm{\pi}} ( {\bm s} ) \eqdef \EE\Big[ \sum_{p=1}^\infty \gamma^p {\cal R}_c^{\bm{\pi}} ( {\bm s}_p ) \!~ | \!~ 
{\bm s}_1 = {\bm s} , \bm{\pi} \Big] \eqs.
\eeq 
To simplify the notation, we define the  vector ${\bm V}^{\bm{\pi}} \in \RR^{|\cal S| } $ 
through stacking up $V^{\bm{\pi}} ( {\bm s} )$  in \eqref{eq:loc_reward} for all ${\bm s}$. 
By definition, $V^{\bm {\pi}}$ satisfies the  Bellman equation  
\beq\label{eq:bellman}
{\bm V}^{\bm{\pi}} = {\bm R}_c^{\bm{\pi}} + \gamma {\bm P}^{\bm{\pi}} {\bm V}^{\bm{\pi}} \eqs,
\eeq
where 
${\bm R}_c^{\bm{\pi}}$ is obtained by stacking up 
\eqref{eq:reward}
and $[ {\bm P}^{\bm{\pi}} ]_{ {\bm s}, {\bm s}' } \eqdef 
\EE_{ \bm{\pi} } [ {\cal P}_{{\bm s}, {\bm s}'}^{ {\bm a} } ]$
is the expected transition matrix.  Moreover, it can be shown that ${\bm V}^{\bm {\pi}}$ is the unique solution of \eqref{eq:bellman}.

When the number of states is large, it is impossible to store ${\bm V}^{\bm {\pi}}$. Instead,   
our goal  is to learn an approximate version of the value function via function approximation.  
In specific,  we  approximate ${V}^{\bm{\pi}}({\bm s})$ using the family of  linear functions 
\begin{align*}
\bigl \{ V_{\prm} ( {\bm s} ) \eqdef \bm{\phi}^\top ( {\bm s} ) \prm \colon \prm \in \RR^d \} , 
\end{align*}
where $\prm \in \RR^d$ is the parameter, $\bm{\phi} ( {\bm s} ) \colon {\cal S} \rightarrow \RR^d$ is a known dictionary consisting of $d$ features, e.g., a feature mapping induced by a neural network. 
To simplify the notation, we 
 define  $\bm{\Phi} \eqdef ( ... ; \bm{\phi}^\top ( {\bm s} ); ... ) \in \RR^{|  {\cal S} | \times d} $ and let ${\bm V} _{\prm} \in \RR^{|\cal S|} $ be the vector constructed by stacking up $\{ V_{\prm}({\bm s}) \}_{{\bm s} \in {\cal S} } $.

With function approximation,  our problem is reduced to finding a
$\prm \in \RR^d$ such that ${\bm V}_{\prm} \approx {\bm V}^{\bm \pi}$.  Specifically, we seek for $\prm$ 
such that the mean squared projected Bellman error (MSPBE) 
\begin{align} \label{eq:mspbe}
{\sf MSPBE}^\star ( \prm )  & \eqdef \frac{1}{2} \Big \| {\bm \Pi}_{\bm{\Phi}} \Bigl  ( 
{\bm V}_{\prm} - \gamma {\bm P}^{\bm{\pi}}  {\bm V}_{\prm}  
- {\bm R}_c^{\bm{\pi}} \Big ) \Big\|_{ {\bm D} }^2  +\rho \| \prm \|^2    
 \end{align}
is minimized,
where ${\bm D} = \text{diag} [ \{ \mu^{\bm \pi}({\bm s}) \}_{{\bm s} \in {\cal S} } ]  \in \RR^{ |{  \cal S} | \times | {\cal S} | } $ is a diagonal matrix constructed using the stationary  
distribution of ${\bm \pi}$, $ {\bm \Pi}_{\bm{\Phi}} \colon \RR^{| {\cal S} | } \rightarrow \RR^{| {\cal S} | } $ is the projection onto subspace $\{ \bm{\Phi} \prm \colon \prm \in \RR^d \}$, 
defined as ${\bm \Pi}_{\bm{\Phi}} = \bm{\Phi} ( \bm{\Phi}^\top {\bm D} \bm{\Phi} )^{-1} \bm{\Phi}^\top {\bm D}$,  
and $\rho \geq 0$ is a free
parameter controlling the regularization on $\prm$.   
For any positive semidefinite matrix ${\bm A} $, we define  $\| {\bm v} \|_{\bm A} = \sqrt{ {\bm v} ^\top {\bm A} {\bm v} } $ for any vector ${\bm v}$. By direct computation, when  $\bm{\Phi}^\top {\bm D} \bm{\Phi}$ is invertible,  the MSPBE defined  in \eqref{eq:mspbe}  can be written as 
\beq\label{eq:population_mspbe}
 {\sf MSPBE}^\star ( \prm )  = \frac{1}{2} \Big\| \bm{\Phi}^\top {\bm D} \Big(
{\bm V}_{\prm} - \gamma {\bm P}^{\bm{\pi}}  {\bm V}_{\prm}  
- {\bm R}_c^{\bm{\pi}} \Big) \Big\|_{ (\bm{\Phi}^\top {\bm D} \bm{\Phi})^{-1} }^2
+ \rho \| \prm \|^2  \eqs =  \frac{1}{2} \Big\| {\bm A} \prm - {\bm b} \Big\|_{ {\bm C}^{-1} }^2 + \rho \| \prm \| ^2 ,
\eeq
where we define 
${\bm A} \eqdef \EE \big[ \bm{\phi} ( {\bm s}_p ) \big( \bm{\phi} ( {\bm s}_p ) - \gamma \bm{\phi} ( {\bm s}_{p+1} ) \big)^\top \big],$ ${\bm C} \eqdef \EE \big[ \bm{\phi} ( {\bm s}_p ) \bm{\phi}^\top ( {\bm s}_p ) \big],$ and $ {\bm b} \eqdef \EE \big[ 
{\cal R}_c^{\bm{\pi}} ( {\bm s}_p ) \bm{\phi} ( {\bm s}_p ) \big]$.  Here  the expectations  in ${\bm A},$ ${\bm b}$, and ${\bm C}$ are all taken with respect to (\wrt) the stationary distribution $\mu^{\bm \pi}$.
Furthermore, when ${\bm A}$  is full rank and ${\bm C}$ is positive definite, it can be shown that the MSPBE in \eqref{eq:population_mspbe} has a unique minimizer.

To obtain a practical optimization problem,  we replace  the expectations above by  their sampled averages from $M$ samples. In specific, for a given policy $\bm{\pi}$,  
a finite state-action sequence 
$\{ {\bm s}_p, {\bm a}_p \}_{p=1}^{M}$ is simulated from the multi-agent 
MDP using  joint policy ${\bm \pi}$. We also observe  ${\bm s}_{M+1}$, the next state of ${\bm s}_{M}$. 
Then we construct the sampled versions of ${\bm A}$, ${\bm b}$, ${\bm C}$, denoted respectively by $\hat{\bm A}$, $\hat{\bm C}$, $\hat{\bm b}$, as
\beq\label{eq:empirical_matrices} 
\begin{split}
& \textstyle  \hat{\bm A} \eqdef 
\frac{1}{M} \sum_{p=1}^{M} {\bm A}_p,~
\hat{\bm C} \eqdef \frac{1}{M} \sum_{p=1}^{M} {\bm C}_p,~
\hat{\bm b} \eqdef \frac{1}{M} \sum_{p=1}^{M} {\bm b}_p,~~\text{with} \\
& 
{\bm A}_p \eqdef \bm{\phi} ( {\bm s}_p ) \big( \bm{\phi} ( {\bm s}_p ) - \gamma \bm{\phi} ( {\bm s}_{p+1} ) \big)^\top,~
{\bm C}_p \eqdef \bm{\phi} ( {\bm s}_p ) \bm{\phi}^\top ( {\bm s}_p ),~
{\bm b}_p \eqdef {\cal R}_c ( {\bm s}_p, {\bm a}_p ) \bm{\phi} ( {\bm s}_p ) \eqs,
\end{split}
\eeq
where 
${\cal R}_c ( {\bm s}_p, {\bm a}_p ) \eqdef N^{-1} \sum_{i=1}^N {\cal R}_i( {\bm s}_p, {\bm a}_p)$ is the average of the local rewards received by each agent when taking action ${\bm a}_p$ at state ${\bm s}_p$. Here we assume that $M$ is sufficiently large such that $\hat {\bm C}$ is invertible and $\hat {\bm A}$ is full rank. Using the terms defined in \eqref{eq:empirical_matrices}, we obtain the empirical MSPBE
\beq\label{eqMSPBE2}
{\sf MSPBE} ( \prm ) \eqdef  \frac{1}{2} \Big\| \hat{\bm A} \prm - \hat{\bm b} \Big\|_{ \hat{\bm C}^{-1} }^2 + \rho \| \prm \| ^2  \eqs,
\eeq
which converges to $  {\sf MSPBE}^\star ( \prm )$ as $M \rightarrow \infty$.
Let $\hat \prm$ be a minimizer of the empirical MSPBE, our estimation of ${\bm V}^{\bm \pi}$ is given by $ {\bm \Phi} \hat \prm$. Since the rewards $ \{ {\cal R}_i ( {\bm s}_p, {\bm a}_p) \}_{i=1}^N $  are  private to each agent, it is impossible for any agent to  compute $ {\cal R}_c ( {\bm s}_p, {\bm a}_p ) $,
and minimize the empirical MSPBE \eqref{eqMSPBE2} independently.  




\textbf{Multi-agent, Primal-dual, Finite-sum Optimization}~~
Recall that under the multi-agent MDP, the agents are able to observe the states and the joint actions, but can only observe their local rewards. Thus, each agent is able to compute $\hat {\bm A}$ and $\hat {\bm C}$ defined in \eqref{eq:empirical_matrices}, but is unable to obtain $\hat {\bm b}$. To resolve this issue, for any $i\in \{ 1, \ldots, N\}$ and any $p \in \{ 1, \ldots, M \}$, we define  
${\bm b}_{p,i} \eqdef {\cal R}_i ( {\bm s}_p, {\bm a}_p ) \bm{\phi} ( {\bm s}_p )$ and $\hat{\bm b}_i \eqdef M^{-1} \sum_{p=1}^{M} {\bm b}_{p,i}$, which  are   known 
to agent $i$ only.
By direct computation, it  is easy to verify that
minimizing ${\sf MSPBE}(\prm)$ in \eqref{eqMSPBE2} is equivalent to solving
\beq \label{eq:prob}
\min_{\prm \in \RR^d} ~\frac{1}{N} \sum_{i=1}^N {\sf MSPBE}_i( \prm ) ~~ \text{where}~~
{\sf MSPBE}_i ( \prm ) \eqdef \frac{1}{2} \Big\| \hat{\bm A} \prm - \hat{\bm b}_i \Big\|_{ \hat{\bm C}^{-1} }^2 + \rho \| \prm \|^2 \eqs.
\eeq
The equivalence  can be seen  by comparing the optimality conditions of  two optimization problems.

Importantly, \eqref{eq:prob} is a
\emph{multi-agent optimization
problems} \citep{nedic2009distributed} whose objective is to minimize a summation of $N$ local functions 
coupled together by the common parameter $\prm$. 
Here ${\sf MSPBE}_i ( \prm ) $ is private to agent $i$ and the parameter $\prm$ is shared by all agents.
As inspired by \citep{nedic2003least, liu2015finite, du2017stochastic}, using Fenchel duality, we obtain  the conjugate form
of ${\sf MSPBE}_i(\prm)$, \ie
\beq\label{eq:fenchel}
\frac{1}{2} \Big\| \hat{\bm A} \prm - \hat{\bm b}_i \Big\|_{ \hat{\bm C}^{-1} }^2 +  \rho \| \prm \|^2   = 
\max_{ {\bm w}_i \in \RR^d } \Big( {\bm w}_i^\top \big( \hat{\bm A} \prm - \hat{\bm b}_i \big)
- \frac{1}{2}  {\bm w}_i^\top \hat{\bm C} {\bm w}_i  \Big) +  \rho \| \prm \|^2  \eqs.
\eeq
Observe that each of $\hat{\bm A}, \hat{\bm C}, \hat{\bm b}_i$ can be expressed as 
a finite sum of matrices/vectors. By \eqref{eq:fenchel}, problem 
\eqref{eq:prob} is equivalent to a 
 \emph{multi-agent}, 
\emph{primal-dual} and \emph{finite-sum} optimization problem:
\beq \label{eq:opt_pd}
\min_{\prm \in \RR^d} \max_{ {\bm w}_i   \in  \RR^d ,i = 1,...,N } \frac{1}{NM} \sum_{i=1}^N \sum_{p=1}^M 
\underbrace{\big( {\bm w}_i^\top {\bm A}_p \prm - {\bm b}_{p,i}^\top {\bm w}_i - \frac{1}{2} {\bm w}_i^\top {\bm C}_p {\bm w}_i + \frac{\rho}{2} \| \prm \|^2 \big)}_{\eqdef J_{i,p} ( \prm, {\bm w}_i )} \eqs.
\eeq
Hereafter, the global objective function is denoted by $J(\prm,  \{ {\bm w}_i \}_{i=1}^N ) 
 \eqdef (1/NM) \sum_{i=1}^N \sum_{p=1}^M J_{i,p} (\prm, {\bm w}_i )$, which is convex  \wrt the primal variable $\prm$ and is concave \wrt the dual variable $\{{\bm w}_i\}_{i=1}^N$.

It is worth noting that  the challenges in solving \eqref{eq:opt_pd} are three-fold. 
First, to obtain a saddle-point solution
$( \{{\bm w}_i\}_{i=1}^N, \prm )$, 
any  algorithm for \eqref{eq:opt_pd} needs  to update the primal and dual variables  simultaneously,
which can be difficult as objective function needs
not be strongly convex with respect to $\prm$. 
In this case,  it is nontrivial to compute a solution efficiently.
Second, the objective function of \eqref{eq:opt_pd} consists
of a sum of $M$ functions, with $M \gg 1$ potentially, such that
conventional primal-dual methods \citep{chambolle2016ergodic} can no longer 
be applied due to the increased complexity. Lastly, 
since $\prm$ is shared by all the agents, 
when solving \eqref{eq:opt_pd}, the $N$ agents need to reach a  consensus on
 $\prm$ without sharing the local functions, e.g.,  
$J_{i,p}(\cdot)$ has to remain unknown
to all agents except for agent $i$ due to privacy concerns. 
Although finite-sum convex optimization problems  with shared variables are well-studied,   new algorithms and theory are  needed for    convex-concave saddle-point problems. 
Next, we propose  a novel decentralized first-order algorithm that tackles these difficulties and converges to a saddle-point solution of \eqref{eq:opt_pd} with linear rate.
\vspace{-.1cm}

\section{Primal-dual Distributed Incremental Aggregated Gradient Method}\vspace{-.1cm}\label{sec:iag}

We are ready to introduce our algorithm for solving  the optimization problem in \eqref{eq:opt_pd}. Since $\prm $ is shared by all the $N$ agents, the agents need to exchange information so as to reach a consensual solution. 
Let us first specify the communication model. 
We assume that the $N$ agents communicate over a network specified by 
a connected and  undirected graph 
$G = (V,E)$, with $V = [N] = \{1,...,N\}$ and $E \subseteq 
V \times V$ being its vertex set and  edge set, respectively. 
Over $G$, it is possible to define a doubly stochastic matrix ${\bm W}$
such that $W_{ij} = 0$ if $(i,j) \notin E$ and ${\bm W} {\bf 1} = {\bm W}^\top {\bf 1} = {\bf 1}$, note $\lambda \eqdef \lambda_{\sf max} ( {\bm W}  - N^{-1} {\bf 1}   {\bf 1} ^\top ) < 1$ since $G$ is connected. 
Notice that the edges in $G$ may be formed independently
of the coupling between agents in the MDP induced by the stochastic policy ${\bm \pi}$.  

We handle problem \eqref{eq:opt_pd} by judiciously combining
the techniques of \emph{dynamic consensus} \citep{qu2017harnessing,zhu2010discrete} 
and \emph{stochastic (or incremental) average gradient} (SAG) \citep{gurbuzbalaban2017convergence,schmidt2017minimizing},
which have been 
developed independently in the control and machine learning 
communities, respectively. 
From a high level viewpoint, our method utilizes a gradient estimator which 
tracks the gradient over \emph{space} (across $N$ agents) 
and \emph{time} (across $M$ samples).
To proceed with our development while explaining the intuitions, 
we first investigate a centralized and batch algorithm
for solving \eqref{eq:opt_pd}.

\textbf{Centralized Primal-dual Optimization}~~
Consider the primal-dual gradient updates. For any $t\geq 1$,  at the $t$-th  iteration, we update the primal and dual variables by   
\beq \label{eq:fgd}
\prm^{t+1} = \prm^t - \gamma_1 \grd_{\prm} J ( \prm^t, \{ {\bm w}_i^t \}_{i=1}^N ), \qquad {\bm w}_i^{t+1} = {\bm w}_i^t +  \gamma_2 \grd_{{\bm w}_i} J( \prm^t, \{ {\bm w}_i^t \}_{i=1}^N ),~i \in [N] \eqs,
\eeq
where $\gamma_1, \gamma_2 > 0$ are step sizes,
which is a simple application of a gradient descent/ascent update to the 
primal/dual variables. As shown by \citet{du2017stochastic},  when $\hat {\bm A}$ is full rank and $\hat {\bm C}$ is invertible, the Jacobian matrix of the primal-dual optimal condition
is full rank. Thus, within a certain
range of step size $(\gamma_1, \gamma_2)$, 
recursion \eqref{eq:fgd} converges 
linearly to the optimal solution of \eqref{eq:opt_pd}.

\textbf{Proposed Method}~~
The primal-dual 
gradient method in \eqref{eq:fgd} serves as a reasonable template 
for developing an efficient   decentralized algorithm 
for \eqref{eq:opt_pd}. 
Let us focus on the update of the primal variable $\prm$ in  \eqref{eq:fgd}, which is a more challenging part since $\prm $ is shared by all $N$ agents. 
To evaluate the gradient \wrt $\prm$, we observe that -- (a) 
agent $i$ does not have access to the functions, $ \{ J_{j,p} (\cdot), j \neq i \} $, 
of the other agents; (b) computing the gradient requires summing up
the contributions from $M$ samples. As $M \gg 1$, 
doing so is undesirable since the computation complexity would be
${\cal O}(Md)$. 

We circumvent the above issues by utilizing a \emph{double
gradient tracking} scheme for the primal $\prm$-update and an incremental  
update scheme for the local dual ${\bm w}_i$-update 
in the following primal-dual distributed incremental aggregated gradient 
({\sf PD-DistIAG}) method. Here each agent $i \in [N]$ maintains a local copy of  the primal parameter $\{ \prm _i^t \} _{t\geq 1}$. 
We construct sequences $\{ {\bm s}_i^t \}_{t\geq 1} $ and $\{ {\bm d}_i^t \}_{t\geq 1}$ to track the gradients  with respect to  $\prm$ and ${\bm w}_i$, respectively. Similar to \eqref{eq:fgd}, in the $t$-th iteration, we update the  dual variable via gradient update  using  ${\bm d}_i^t$. As for the primal variable, to achieve consensus, each $\prm_i^{t+1} $ is obtained  by first combining    $\{ \prm_i^t \} _{i \in [N]}$ using the weight matrix ${\bm W}$, and then update in the direction of ${\bm s}_i^t$.    The details of our method are presented in Algorithm \ref{algo:main}.

\begin{algorithm}[t]
\caption{\textbf{{\sf PD-DistIAG} Method} for Multi-agent, Primal-dual, Finite-sum Optimization} 
\label{algo:main}
\begin{algorithmic}
\STATE{{ \bf Input}: Initial estimators $\{ \prm_i^1, {\bm w}_i^1  \}_{i \in [N]}$,  initial gradient estimators ${\bm s}_i^0 = {\bm d}_i^0 =  {\bf 0}$, $\forall~i \in [N]$, 
initial counter $\tau_p^0 = 0$, $\forall~p \in [M]$,  and stepsizes $\gamma_1, \gamma _2 > 0$. }
\FOR {$t\geq 1$}
\STATE{The agents pick a common sample indexed by 
$p_t \in \{1,...,M\}$.}
\STATE{Update the counter variable as:\vspace{-.3cm}
\beq \label{eq:tau_upd}
\tau_{p_t}^t = t,~~\tau_p^t = \tau_p^{t-1},~\forall~p \neq p_t \eqs.\vspace{-.4cm}
\eeq}
\FOR{each agent $i \in \{ 1, \ldots, N\} $}
\STATE{Update the gradient surrogates by\vspace{-.2cm}
\begin{align} 
{\bm s}_i^t & \textstyle = \sum_{j=1}^N W_{ij} {\bm s}_j^{t-1} + \frac{1}{M} \Big[  \grd_{\prm} J_{i,p_t} ( \prm_i^t, {\bm w}_i^t ) - 
\grd_{\prm} J_{i,p_t} ( \prm_i^{\tau_{p_t}^{t-1}}, {\bm w}_i^{\tau_{p_t}^{t-1}} ) 
\Big] \eqs, \label{eq:s_upd}\\
{\bm d}_i^t & \textstyle = {\bm d}_i^{t-1} + \frac{1}{M} 
 \Big[  \grd_{\bm w_i} J_{i,p_t} ( \prm_i^t, {\bm w}_i^t ) - \grd_{\bm w_i} J_{i,p_t} ( \prm_i^{\tau_{p_t}^{t-1}}, {\bm w}_i^{\tau_{p_t}^{t-1}} ) \Big] \eqs, \label{eq:d_upd}
\end{align}
where $\grd_{\prm} J_{i,p} ( \prm_i^{0}, {\bm w}_i^{0} ) = {\bm 0} $
and $\grd_{{\bm w}_i} J_{i,p} ( \prm_i^{0}, {\bm w}_i^{0} ) = {\bm 0} $ 
for all $p \in [M]$ for initialization.}\vspace{.1cm}
\STATE{Perform primal-dual updates using ${\bm s}_i^t, {\bm d}_i^t$ as surrogates for the gradients \wrt $\prm$
and ${\bm w}_i$:
\beq\textstyle  \label{eq:pd_alg}
\prm_i^{t+1} = \sum_{j=1}^N W_{ij} \prm_j^t - \gamma_1 {\bm s}_i^t,~~
{\bm w}_i^{t+1} = {\bm w}_i^t + \gamma_2 {\bm d}_i^t \eqs.\vspace{-.1cm}
\eeq
} 
\ENDFOR
\ENDFOR 
\end{algorithmic} 
\end{algorithm}

Let us explain the intuition behind the {\sf PD-DistIAG} method
through studying the update \eqref{eq:s_upd}.
Recall that the global gradient desired at iteration $t$ is given by
$\grd_{\prm} J ( \prm^t, \{ {\bm w}_i^t \}_{i=1}^N )$, which represents
a double average -- one over space (across agents) and 
one over time (across samples). 
Now in the case of \eqref{eq:s_upd}, 
the first summand on the right hand side computes
a local average among the neighbors of agent $i$, and thereby tracking 
the global gradient over \emph{space}. This is in fact akin to the 
\emph{gradient tracking} technique in the context of 
distributed optimization \citep{qu2017harnessing}. 
The remaining terms on the right hand side of \eqref{eq:s_upd}
utilize an incremental update rule  akin to 
the SAG method \citep{schmidt2017minimizing}, involving 
a swap-in swap-out operation for the gradients. This 
achieves tracking of the global gradient over \emph{time}.

To gain insights on 
why the scheme works, we note that 
 ${\bm s}_{i}^{t}$ and ${\bm d}^t_{i}$ represent some surrogate functions for the primal and dual gradients. 
Moreover, for the counter variable, using \eqref{eq:tau_upd} we can 
alternatively represent it as 
$\tau_p^t = \max \{ \ell \geq 0 \!~:\!~ \ell \leq t,~ p_\ell = p \}$.
In other words, $\tau_p^t$ is
the iteration index where the $p$-th sample is last visited by the agents
prior to iteration $t$, and if the $p$-th sample has never been visited,
we have $\tau_p^t = 0$.
For any $t\geq 1$, define
${\bm g}_{\prm}(t) \eqdef (1/N)\sum_{i=1}^N {\bm s}_i^t$. 
The following lemma shows that ${\bm g}_{\prm}(t)$ is 
a double average of the primal gradient -- 
it averages over the local gradients across the agents, 
and for each local gradient; it also averages over the past gradients for all the samples 
evaluated up till iteration $t+1$. 
This shows that the average over network for $\{ {\bm s}_i^t \}_{i=1}^N$ 
can always track the double average of the local and past gradients, 
\ie the gradient estimate 
${\bm g}_{\prm}(t) $ is `unbiased' with respect to the network-wide average. 

\begin{Lemma} \label{lem:avg}
For all $t \geq 1$ and consider Algorithm~\ref{algo:main}, it holds that\vspace{-.1cm}
\beq \label{eq:track} \textstyle
{\bm g}_{\prm}(t)  = \frac{1}{NM} \sum_{i=1}^N
\sum_{p=1}^M \grd_{\prm} J_{i,p} ( \prm_i^{\tau_p^t}, {\bm w}_i^{\tau_p^t} ) \eqs.\vspace{-.2cm}
\eeq
\end{Lemma}

\textbf{Proof.} We shall prove the statement using induction. For 
the base case with $t = 1$, 
using \eqref{eq:s_upd} and the update rule specified
in the algorithm, we have\vspace{-.2cm}
\beq \label{eq:base_case}
{\bm g}_{\prm}(1) = \frac{1}{N} \sum_{i=1}^N \frac{1}{M} 
\grd_{\prm} J_{i,p_1} ( \prm_i^1, {\bm w}_i^1 )
= \frac{1}{NM} \sum_{i=1}^N \sum_{p=1}^M 
\grd_{\prm} J_{i,p_t} ( \prm_i^{\tau_p^1}, {\bm w}_i^{\tau_p^1} ) \eqs,\vspace{-.1cm}
\eeq
where we use the fact    
$\grd_{\prm} J_{i,p} ( \prm_i^{\tau_p^1}, {\bm w}_i^{\tau_p^1} ) = \grd_{\prm} J_{i,p} ( \prm_i^{0}, {\bm w}_i^{0} ) = {\bm 0} $
for all $p \neq p_1$
in the above. 
For the induction step, suppose 
\eqref{eq:track} holds up to iteration $t$.
Since ${\bm W}$ is  doubly stochastic, \eqref{eq:s_upd} implies  
\beq  \label{eq:induction_step}
\begin{split}
{\bm g}_{\prm}(t+1) & = 
\frac{1}{N}  \sum_{i=1}^N \bigg \{ \sum_{j=1}^N W_{ij} {\bm s}_j^t 
+ \frac{1}{M} \Big[  \grd_{\prm} J_{i,p_{t+1}} ( \prm_i^{t+1}, {\bm w}_i^{t+1} ) - 
\grd_{\prm} J_{i,p_{t+1}} ( \prm_i^{\tau_{p_{t+1}}^{t}}, {\bm w}_i^{\tau_{p_{t+1}}^{t}} )  \Big]  \bigg\}  
 \\
 & = {\bm g}_{\prm}(t)  +
 \frac{1}{NM} \sum_{i=1}^N \Big[  \grd_{\prm} J_{i,p_{t+1}} ( \prm_i^{t+1}, {\bm w}_i^{t+1} ) - 
\grd_{\prm} J_{i,p_{t+1}} ( \prm_i^{\tau_{p_{t+1}}^{t}}, {\bm w}_i^{\tau_{p_{t+1}}^{t}} )  \Big]\eqs.\\[-.4cm]
\end{split}\vspace{-.4cm}
\eeq
Notice that we have $\tau_{p_{t+1}}^{t+1} = t+1$ 
and $\tau_{p}^{t+1} = \tau_{p}^{t}$ for all $p \neq p_{t+1}$. 
The induction assumption in \eqref{eq:track} can be written as 
\begin{align}  \label{eq:induction_step2} 
{\bm g}_{\prm}(t) =
\frac{1}{NM} \sum_{i=1}^N  \biggl [ \sum_{p\neq p_{t+1} }     \grd_{\prm} J_{i,p} ( \prm_i^{\tau_p^{t+1}}, {\bm w}_i^{\tau_p^{t+1}} )  \biggr ]  + \frac{1}{NM}\sum_{i=1}^N \grd_{\prm} J_{i,p_{t+1} } ( \prm_i^{\tau_{p_{t+1}} ^{t}}, {\bm w}_i^{\tau_{p_{t+1}} ^{t } } )  \eqs. 
\end{align}
Finally, combining  \eqref{eq:induction_step} and \eqref{eq:induction_step2}, 
we obtain the desired result that \eqref{eq:track} holds for the $t+1$th iteration. 
This, together with  \eqref{eq:base_case}, 
establishes Lemma \ref{lem:avg}.
 \hfill \textbf{Q.E.D.}\vspace{.1cm}

As for the dual update \eqref{eq:d_upd},
we observe the variable ${\bm w}_i$ is local to agent $i$.
Therefore its gradient surrogate, ${\bm d}_i^t$, involves only the tracking
step over \emph{time}
[cf.~\eqref{eq:d_upd}], \ie it only averages the gradient over samples.
Combining with Lemma~\ref{lem:avg}
shows that the {\sf PD-DistIAG} method uses gradient surrogates
that are averages over samples despite the disparities across agents.  
Since the average over samples are done in a similar spirit
as the {\sf SAG} method, the proposed method is expected
to converge linearly.

\textbf{Storage and Computation Complexities}~~Let us comment on 
the computational and storage
complexity of {\sf PD-DistIAG} method.
First of all,  
since the method requires accessing the previously evaluated gradients, 
each agent has to store $2M$ such vectors in the memory to avoid
re-evaluating these gradients. Each agent   needs to store
a total of $2Md$ real numbers. 
On the other hand, the per-iteration computation complexity for each agent is 
only ${\cal O}(d)$ as each iteration only requires to evaluate 
the gradient over one sample, as
delineated in \eqref{eq:d_upd}--\eqref{eq:pd_alg}. 


\textbf{Communication Overhead}~~
The {\sf PD-DistIAG} method described in Algorithm \ref{algo:main} requires an information exchange round 
[of ${\bm s}_i^t$ and $\prm_i^t$] among the agents
at every iteration. From an implementation stand point, this may incur 
significant communication overhead when $d \gg 1$, and it is 
especially ineffective 
when the progress made in successive updates of the algorithm
is not significant. 
A natural remedy is to perform multiple \emph{local} updates at the agent 
using different 
samples \emph{without} exchanging information with the neighbors.
In this way, the communication overhead can be reduced. 
Actually, this modification to the {\sf PD-DistIAG} method can 
be generally described using a time varying weight matrix ${\bm W}(t)$,
such that we have ${\bm W}(t) = {\bm I}$ for most of the iteration. 
The convergence of {\sf PD-DistIAG} method in this scenario is part of the 
future work.\vspace{-.1cm}

\subsection{Convergence Analysis}
The {\sf PD-DistIAG} method is built using the techniques of 
(a) primal-dual batch gradient 
descent, (b) gradient tracking for distributed optimization and (c) 
stochastic average gradient, 
where each of them has been independently shown to attain
linear convergence under certain conditions; see \citep{qu2017harnessing,
schmidt2017minimizing,gurbuzbalaban2017convergence,du2017stochastic}. 
Naturally, the {\sf PD-DistIAG} method is also anticipated to converge at
a linear rate. 

To see this, let us consider the condition for the sample selection rule 
of {\sf PD-DistIAG}:\vspace{-.1cm}
 \begin{assumption} \label{ass:bd}
 A sample is selected at least once for every $M$ iterations, $| t - \tau_p^t | \leq M$ for all $p \in [M]$, $t \geq 1$.\vspace{-.1cm}
 \end{assumption}
The assumption requires that every samples 
are visited infinitely often. For example, this can be enforced by
using a cyclical selection rule, \ie $p_t = (t~{\rm mod}~M) + 1$;
or a random sampling scheme \emph{without replacement} 
(\ie random shuffling)
from the pool of $M$ samples.
Finally, it is possible to relax the assumption such that a sample 
can be selected once for every $K$ iterations only, with $K \geq M$. 
The present assumption is made solely for the purpose of ease
of presentation. Moreover,
to ensure that the solution to \eqref{eq:opt_pd} is unique, we consider:\vspace{-.1cm}
 \begin{assumption} \label{ass:fr}
The sampled correlation matrix $\hat{\bm A}$ is full rank, and the sampled covariance $\hat{\bm C}$ is non-singular.\vspace{-.1cm}
 \end{assumption}
The following theorem confirms the linear convergence of {\sf PD-DistIAG}:
\begin{Theorem} \label{thm:main}
Under  A\ref{ass:bd} and  A\ref{ass:fr}, we denote  by $( \prm^\star, \{ {\bm w}_i^\star \}_{i=1}^N )$ the  
primal-dual optimal solution to the optimization problem in \eqref{eq:opt_pd}. Set the 
step sizes as $\gamma_2 = \beta \gamma_1 $ with
$\beta \eqdef 8 ( \rho + \lambda_{\sf max} ( \hat{\bm A}^\top \hat{\bm C}^{-1} \hat{\bm A}) ) / \lambda_{\sf min}( \hat{\bm C})$. 
Define $\overline{\prm}(t) \eqdef \frac{1}{N} \sum_{i=1}^N \prm_i^t$ as 
the average of parameters. 
If the primal step size $\gamma_1$ is sufficiently small, then 
there exists a constant $0 < \sigma < 1$ that 
\beq \label{eq:converge} \notag
\textstyle
\big\| \overline{\prm}(t) - \prm^\star \big\|^2 + (1/\beta N) \sum_{i=1}^N \big\| {\bm w}_i^t - {\bm w}_i^\star \big\|^2 = {\cal O}( \sigma^t ),~~  (1/N) \sum_{i=1}^N \big\| \prm_i^t - \overline{\prm}(t) \big\| = {\cal O}( \sigma^t ) \eqs.
\eeq
If $N,M \gg 1$ and the graph is geometric, a sufficient condition for convergence
is to set $\gamma = {\cal O}(1/ \max\{N^{2},M^2\})$ and the resultant rate 
is $\sigma = 1 - {\cal O}( 1/ \max\{M N^{2}, M^3\})$. 
\end{Theorem}

The result above shows the desirable convergence properties for {\sf PD-DistIAG} 
method -- 
the primal dual solution 
$( \overline{\prm}(t), \{ {\bm w}_i^t \}_{i=1}^N )$ 
converges to $( \prm^\star, \{ {\bm w}_i^\star \}_{i=1}^N )$
at a linear rate; also, the 
consensual error of the local
parameters $\bar{\prm}_i^t$ converges to zero linearly. 
A distinguishing feature of our analysis is that it handles the \emph{worst case}
convergence of the proposed method, rather than the \emph{expected} 
convergence rate popular for stochastic / incremental gradient methods. 


\textbf{Proof Sketch}~~Our proof is   divided  into three steps.  
The first step studies the progress made by the algorithm in 
one iteration, taking into account the non-idealities due to imperfect tracking
of the gradient over space and time. 
This leads to the characterization of a  \emph{Lyapunov vector}. 
The second step analyzes the \emph{coupled} system of one iteration progress
made by  the  Lyapunov vector.
An interesting feature of it is that it 
consists of a series of independently \emph{delayed} terms in the Lyapunov vector. The latter is resulted from the incremental update 
schemes employed in the method.
Here, we study a sufficient condition for the coupled and delayed system to
converge linearly. 
The last step is to derive condition on the step size $\gamma_1$
where the sufficient convergence condition is satisfied. 

Specifically, we study the progress of 
the Lyapunov functions:
\beq \notag 
\begin{array}{c}
\| \widehat{\underline{\bm v}}(t) \|^2  \eqdef \Theta \Big( \big\| \overline{\prm}(t) - \prm^\star \big\|^2 + (1/\beta N) \sum_{i=1}^N \big\| {\bm w}_i^t - {\bm w}_i^\star \big\|^2 \Big),~~ {\cal E}_c(t) \eqdef \frac{1}{N} \sqrt{\sum_{i=1}^N \| \prm_i^t - \overline{\prm}(t) \|^2}, \vspace{.1cm} \\
\textstyle {\cal E}_g(t) \eqdef \frac{1}{N} \sqrt{ \sum_{i=1}^N \big\| {\bm s}_i^t - \frac{1}{NM}    \sum_{j=1}^N \sum_{p=1}^M \grd_{\prm} J_{j,p} ( \prm_j^{\tau_p^t}, {\bm w}_j^{\tau_p^t} ) \big\|^2} \eqs.
\end{array}
\eeq
That is,  $\widehat{\underline{\bm v}}(t)$ is 
a vector whose squared norm is equivalent to a weighted distance to the optimal 
primal-dual solution,  $  {\cal E}_c(t)$ and ${\cal E}_g(t) $ 
  are respectively the consensus errors of the primal parameter
and of the primal \emph{aggregated} gradient. 
These functions form a non-negative 
vector which evolves as:
\beq \label{eq:fin_sys_paper}
\left( 
\begin{array}{c}
\| \widehat{\underline{\bm v}}(t+1) \| \\[.1cm]
{\cal E}_c(t+1) \\[.1cm]
{\cal E}_g(t+1) 
\end{array}
\right)
\leq 
{\bm Q} (\gamma_1)
\left( 
\begin{array}{c}
\max_{ (t-2M)_+ \leq q \leq t }  \| \widehat{\underline{\bm v}}(q) \| \\[.1cm]
\max_{ (t-2M)_+ \leq q \leq t }  {\cal E}_c(q) \\[.1cm]
\max_{ (t-2M)_+ \leq q \leq t }  {\cal E}_g(q) 
\end{array}
\right) \eqs,
\eeq 
where   the matrix ${\bm Q} ( \gamma_1 ) \in \RR^{3 \times 3}$  is defined by (exact form given in the supplementary material)
\beq \label{eq:Q_gamma_mat}
{\bm Q} ( \gamma_1 ) = \left( 
\begin{array}{ccc}
1 - \gamma_1 a_0 + \gamma_1^2 a_1 & \gamma_1 a_2 & 0 \\
0 & \lambda & \gamma_1 \\
\gamma_1 a_3 & a_4 + \gamma_1 a_5 & \lambda + \gamma_1 a_6 
\end{array}
\right) \eqs.
\eeq
In the above, $\lambda \eqdef  \lambda_{\sf max} ( {\bm W} - (1/N) {\bf 1}{\bf 1}^\top ) < 1$,  and
$a_0, ..., a_6$ are some non-negative constants
that depends on
the problem parameters $N$, $M$, the spectral
properties of ${\bm A}$, ${\bm C}$, etc., with $a_0$ being positive.
If we focus only on the first row of the inequality system, we obtain  
\begin{align*}
\| \widehat{\underline{\bm v}}(t + 1) \|
\leq \big( 1 - \gamma_1 a_0 + \gamma_1^2 a_1 ) 
\max_{ (t-2M)_+ \leq q \leq t }  \| \widehat{\underline{\bm v}}(q) \|
+ \gamma_1 a_2 \max_{ (t-2M)_+ \leq q \leq t }  {\cal E}_c(q) \eqs. 
\end{align*}
In fact, when the contribution from ${\cal E}_c(q)$ can be ignored, then 
applying \citep[Lemma 3]{feyzmahdavian2014delayed} shows 
that $\| \widehat{\underline{\bm v}}(t + 1) \|$ converges linearly if
$- \gamma_1 a_0 + \gamma_1^2 a_1 < 0$, which is possible 
as $a_0 > 0$. 
Therefore, if ${\cal E}_c(t)$ also converges linearly, then 
it is anticipated that   ${\cal E}_g(t)$ would do so as well.
In other words, the linear convergence of $\| \widehat{\underline{\bm v}}(t) \|$,
${\cal E}_c(t)$ and ${\cal E}_g(t)$ are all coupled in the inequality system \eqref{eq:fin_sys_paper}. 

Formalizing the above observations, 
Lemma~1 in the supplementary material shows 
a sufficient condition on $\gamma_1$
for linear convergence. 
Specifically, 
if  there exists $\gamma_1>0$ such that  the spectral radius of ${\bm Q}( \gamma_1 )$  in \eqref{eq:Q_gamma_mat} is strictly less than one, 
then each of the Lyapunov functions, $\| \widehat{\underline{\bm v}}(t) \|$, 
${\cal E}_c(t)$, ${\cal E}_g(t)$, would enjoy linear convergence. 
Furthermore, Lemma~2 in the supplementary material gives
an existence proof for such an $\gamma_1$ to exist. 
This concludes the proof.

\textbf{Remark}~~While delayed inequality system
has been studied in \citep{feyzmahdavian2014delayed,gurbuzbalaban2017convergence}
for optimization algorithms,  
the coupled system in \eqref{eq:fin_sys_paper}
is a non-trivial generalization of the above. Importantly, the challenge
here is due to the asymmetry of the system matrix ${\bm Q}$
and the maximum over the past sequences on the right hand side 
are taken \emph{independently}.
To the best of our knowledge, our result is 
the first to characterize the (linear) convergence of such 
coupled and delayed system of inequalities. 
%

\textbf{Extension}~~
Our analysis and algorithm may in fact be applied to solve general problems 
that involves multi-agent and finite-sum optimization, e.g., 
\beq \label{eq:s_opt}
\textstyle 
\min_{ \prm \in \RR^d } ~J(\prm) \eqdef \frac{1}{ NM}\sum_{i=1}^N \sum_{p=1}^M J_{i,p} ( \prm ) \eqs.
\eeq
For instance, these problems may arise in 
multi-agent empirical risk minimization, where data samples are kept independently
by agents. 
Our analysis, especially with convergence for
inequality systems of the form \eqref{eq:fin_sys_paper},
can be applied to study a similar double averaging algorithm with just
the primal variable. 
In particular, we only require the sum function $J(\prm)$
to be strongly convex, and 
the objective functions $J_{i,p}(\cdot)$ to be smooth in order
to achieve linear convergence. 
We believe that such extension is of independent interest to the community. 
At the time of submission, a recent work \citep{pu2018distributed} 
applied a related double averaging
distributed algorithm 
to a \emph{stochastic version} of \eqref{eq:s_opt}. 
However, their  convergence rate is sub-linear
as they considered a stochastic optimization setting.


\section{Numerical Experiments}\label{sec:ne}
To verify the performance of our proposed method, we conduct 
an experiment 
on the \texttt{mountaincar} dataset \citep{sutton1998reinforcement}
under a setting similar to \citep{du2017stochastic} -- 
to collect the dataset, we ran Sarsa with $d=300$ features to obtain the policy, 
then we generate the trajectories of actions and states according to the policy
with $M$ samples.
For each sample $p$, we generate the local reward, $R_i( s_{p,i}, a_{p,i} )$ 
by assigning a random 
portion for the reward to each agent such that the average
of the local rewards equals ${\cal R}_c ( {\bm s}_p, {\bm a}_p )$. 

We compare our method to several centralized methods -- 
{\sf PDBG} is the primal-dual gradient descent method in \eqref{eq:fgd}, 
{\sf GTD2} \citep{sutton2009convergent}, and {\sf SAGA} \citep{du2017stochastic}.
Notably, {\sf SAGA} has linear convergence while only requiring 
an incremental update step of low complexity.
For {\sf PD-DistIAG}, we simulate a communication network 
with $N=10$ agents, connected on an Erdos-Renyi graph 
generated with connectivity of $0.2$; for the step sizes,
we set $\gamma_1 = 0.005 / \lambda_{\sf max}( \hat{\bm A} )$,
$\gamma_2 = 5 \times 10^{-3}$. 

\begin{figure}[h]
\centering
\includegraphics[width=.32\linewidth]{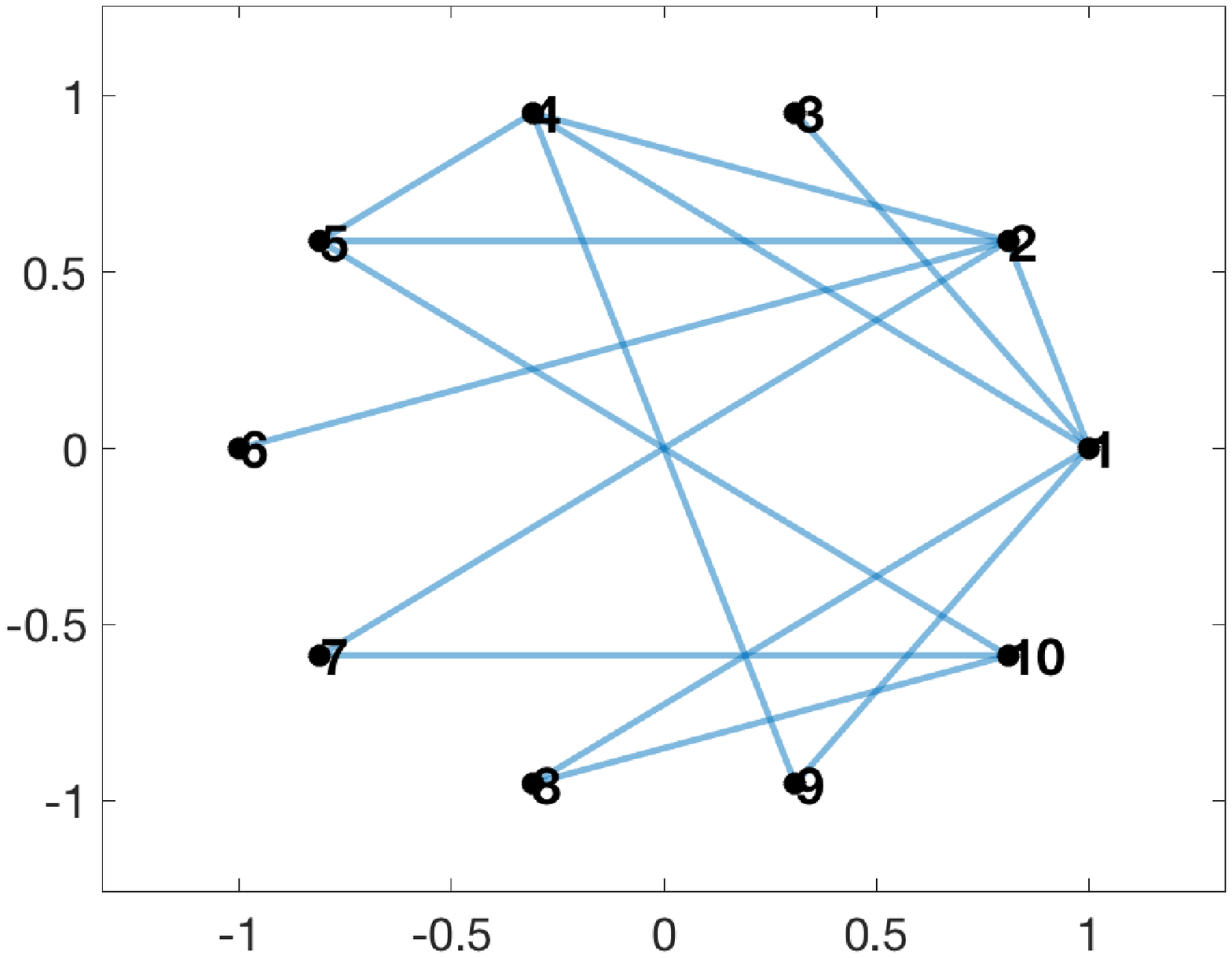}~
\includegraphics[width=.32\linewidth]{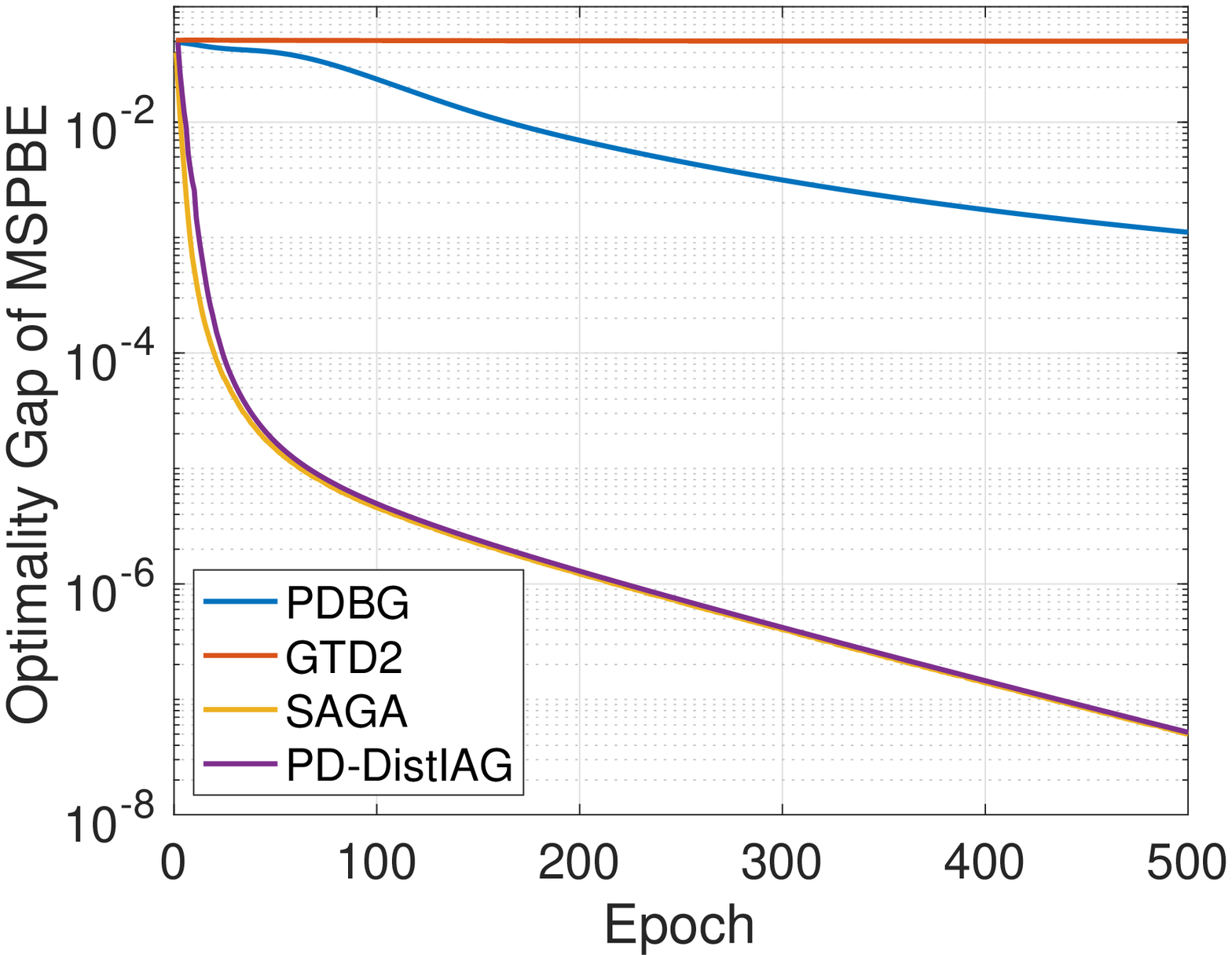}~
\includegraphics[width=.32\linewidth]{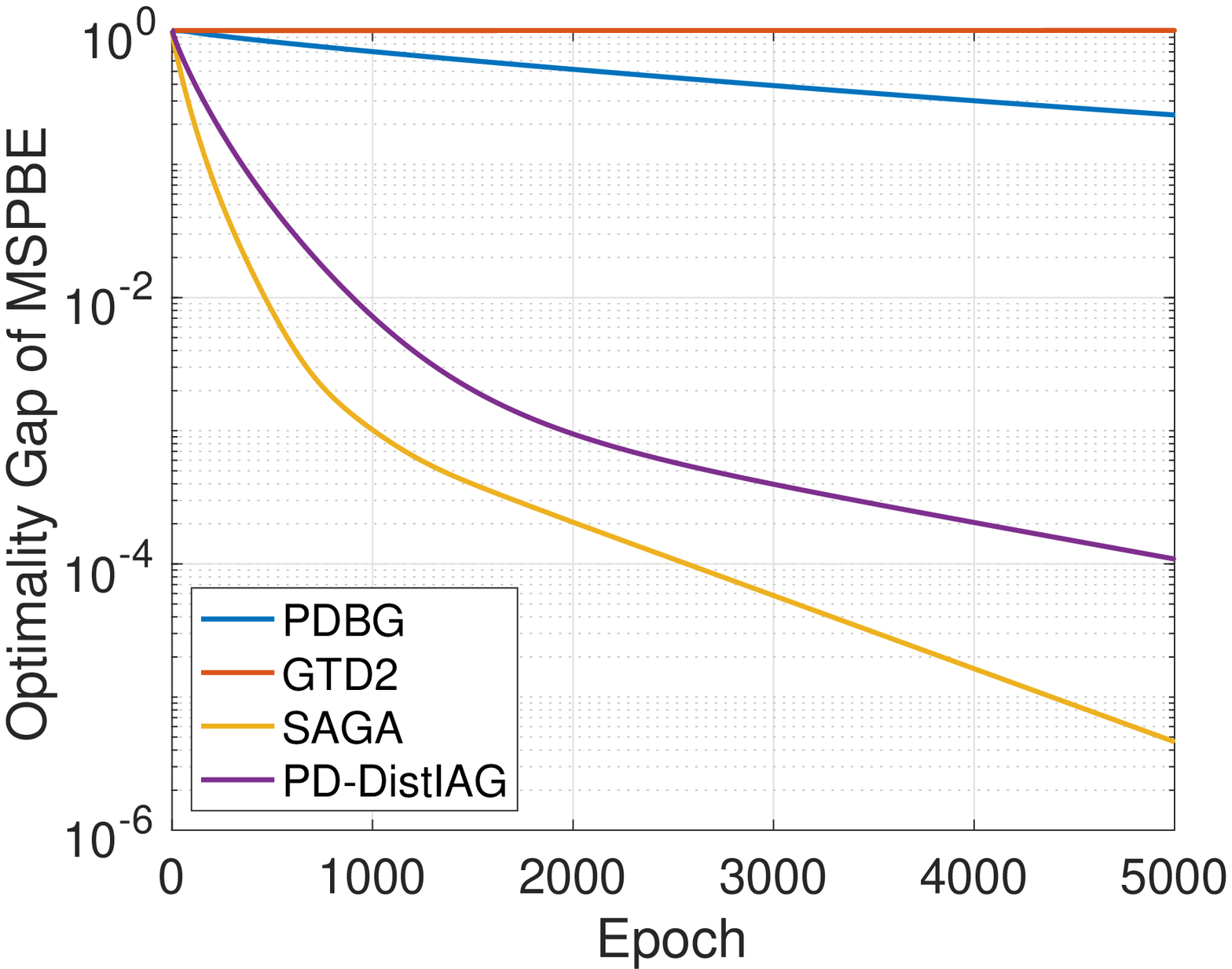}
\caption{Experiment with \texttt{mountaincar} dataset. For this problem, we have  $d=300$, $M=5000$ samples, and there are $N=10$ agents. 
(Left) Graph Topology. (Middle) $\rho = 0.01$. (Right) $\rho = 0$.} \label{fig:one}
\end{figure}

Figure~\ref{fig:one} compares the optimality gap in terms of MSPBE 
of different algorithms 
against the epoch number, defined as $(t / M)$.  
For {\sf PD-DistIAG}, we compare its optimality gap 
in MSPBE as the average objective, \ie it is
$(1/N) \sum_{i=1}^N {\sf MSPBE} ( \prm_i^t ) - {\sf MSPBE}( \prm^\star )$. 
As seen in the left panel, when the regularization factor is high with 
$\rho > 0$, the convergence speed of {\sf PD-DistIAG} 
is comparable to that of {\sf SAGA}; meanwhile with $\rho = 0$, 
the {\sf PD-DistIAG} converges at a slower speed than {\sf SAGA}.
Nevertheless, in both cases, the {\sf PD-DistIAG} method converges 
faster than the other methods except for {\sf SAGA}. 
Additional experiments are presented in the supplementary materials
to compare the performance at different topology and regularization parameter.


\textbf{Conclusion}~~
In this paper, we have studied the policy evaluation problem in \emph{multi-agent}
reinforcement learning. Utilizing Fenchel duality, a double averaging scheme is proposed to 
tackle the primal-dual, multi-agent, and finite-sum optimization arises.
The proposed {\sf PD-DistIAG} method demonstrates
linear convergence under reasonable assumptions. 

%

\numberwithin{equation}{section}

\renewcommand\thefigure{\thesection.\arabic{figure}}    
\setcounter{figure}{0}


\appendix
\section{Proof of Theorem~1}
We repeat the statement of the theorem as follows:

 
\begin{Theorem*} \label{thm:main_app}
Under  A\ref{ass:bd} and  A\ref{ass:fr}, we denote  by $( \prm^\star, \{ {\bm w}_i^\star \}_{i=1}^N )$ the  
primal-dual optimal solution to the optimization problem in \eqref{eq:opt_pd}. Set the 
step sizes as $\gamma_2 = \beta \gamma_1 $ with
$\beta \eqdef 8 ( \rho + \lambda_{\sf max} ( \hat{\bm A}^\top \hat{\bm C}^{-1} \hat{\bm A}) ) / \lambda_{\sf min}( \hat{\bm C})$. 
Define $\overline{\prm}(t) \eqdef \frac{1}{N} \sum_{i=1}^N \prm_i^t$ as 
the average of parameters. 
If the primal step size $\gamma_1$ is sufficiently small, then 
there exists a constant $0 < \sigma < 1$ that 
\beq \label{eq:converge} \notag
\textstyle
\big\| \overline{\prm}(t) - \prm^\star \big\|^2 + (1/\beta N) \sum_{i=1}^N \big\| {\bm w}_i^t - {\bm w}_i^\star \big\|^2 = {\cal O}( \sigma^t ),~~  (1/N) \sum_{i=1}^N \big\| \prm_i^t - \overline{\prm}(t) \big\| = {\cal O}( \sigma^t ) \eqs.
\eeq
If $N,M \gg 1$ and the graph is geometric with $\lambda = 1 - c/N$ for  $c>0$, a sufficient condition for convergence
is to set $\gamma = {\cal O}(1/ \max\{N^2,M^2\})$ and the resultant rate 
is $\sigma = 1 - {\cal O}( 1/ \max\{M N^2, M^3\})$. 
\end{Theorem*}

\paragraph{Notation}
We first define a set of notations pertaining to the proof. 
For any $\beta > 0$, observe that the primal-dual optimal solution,
$(  \prm^\star, \{ {\bm w}_i ^{\star}\}_{i=1}^N )$,  
to the optimization problem  \eqref{eq:opt_pd} can be written as 
\beq \label{eq:opt_cond}
\underbrace{\left( 
\begin{array}{cccc}
\rho {\bm I} & \sqrt{\frac{\beta}{N}} \hat{\bm A}^\top & \cdots & \sqrt{\frac{\beta}{N}} \hat{\bm A}^\top \\
-\sqrt{\frac{\beta}{N}} \hat{\bm A} & \beta \hat{\bm C} & \cdots & \cdots \\
\vdots & {\bm 0} & \ddots & {\bm 0} \\
-\sqrt{\frac{\beta}{N}} \hat{\bm A} & \cdots & \cdots & \beta \hat{\bm C} 
\end{array}
\right)}_{\eqdef {\bm G}}
\left( \begin{array}{c}
\prm^\star \\
\frac{1}{\sqrt{\beta N}}{\bm w}_1^\star \\
\vdots \\
\frac{1}{\sqrt{\beta N}}{\bm w}_N^\star 
\end{array}
\right) 
= 
\left( \begin{array}{c}
{\bm 0} \\
-\sqrt{\frac{\beta}{N}}{\bm b}_1 \\
\vdots \\
-\sqrt{\frac{\beta}{N}}{\bm b}_N 
\end{array}
\right) \eqs, 
\eeq
where we  denote the matrix on the left hand side as ${\bm G}$.  This equation can be obtained by checking the first-order optimality condition. In addition, for any $p \in \{ 1, \ldots, M \}$, we  define the  ${\bm G}_p$ as 
\beq\label{eq:matrix_Gp}
{\bm G}_p \eqdef \left( 
\begin{array}{cccc}
\rho {\bm I} & \sqrt{\frac{\beta}{N}} {\bm A}_p^\top & \cdots & \sqrt{\frac{\beta}{N}} {\bm A}^\top \\
-\sqrt{\frac{\beta}{N}}{\bm A}_p & \beta {\bm C}_p & \cdots & \cdots \\
\vdots & {\bm 0} & \ddots & {\bm 0} \\
-\sqrt{\frac{\beta}{N}}{\bm A}_p & \cdots & \cdots & \beta {\bm C}_p
\end{array}
\right) \eqs.
\eeq 
By definition, ${\bm G}$ is the sample average of $ \{ {\bm G}_p \}_{p=1}^M$.
Define $\bar{\prm}(t) \eqdef (1/N) \sum_{i=1}^N \prm_i^t$ as the average of the local parameters at iteration $t$. Furthermore, we define
\begin{align}
	{\bm h}_{\prm}(t)  \eqdef \rho \bar{\prm}(t) + \frac{1}{N} \sum_{i=1}^N \hat{\bm A}^\top {\bm w}_i^t, & \qquad {\bm g}_{\prm}(t) \eqdef \frac{1}{NM} \sum_{i=1}^N \sum_{p=1}^M \big( \rho \prm_i^{\tau_p^t} + {\bm A}_p^\top {\bm w}_i^{\tau_p^t} \big)  \eqs, \label{eq:4terms_1}  \\
	{\bm h}_{ {\bm w}_i } (t) \eqdef \hat{\bm A} \bar{\prm}(t) - \hat{\bm C} {\bm w}_i^t - \hat{\bm b}_i, & \qquad 
	{\bm g}_{ {\bm w}_i } (t) \eqdef \frac{1}{M} \sum_{p=1}^M \big( {\bm A}_p \prm_i^{\tau_p^t} - {\bm C}_p {\bm w}_i^{\tau_p^t} - {\bm b}_{p,i} \big) \eqs, \label{eq:4terms_2}
\end{align}
where ${\bm h}_{\prm}(t)$ and $ {\bm h}_{\bm w}(t) : = [{\bm h}_{ {\bm w}_1} (t), \cdots, 	{\bm h}_{ {\bm w}_N } (t)]$ represent the gradients evaluated by a \emph{centralized} and 
\emph{batch} algorithm. 
Note that ${\bm g}_{\prm}(t)$ defined in \eqref{eq:4terms_1} coincides with that in \eqref{eq:track}.
Using Lemma~\ref{lem:avg}, 
it can be checked that $\bar{\prm}(t+1) = \bar{\prm}(t) - \gamma_1 {\bm g}_{\prm}(t)$
and ${\bm w}_i^{t+1} = {\bm w}_i^t - \gamma_2 {\bm g}_{ {\bm w}_i } (t)$
for all $t \geq 1$.
That is, $\bar{\prm}(t+1)$ and ${\bm w}_i^{t+1} $ can be viewed as primal-dual updates using  ${\bm g}_{\prm}(t)$ and ${\bm g}_{ {\bm w}_i } (t)$, which are decentralized counterparts of gradients ${\bm h}_{\prm}(t)$ and ${\bm h}_{ {\bm w}_i} (t)$ defined in \eqref{eq:4terms_1} \eqref{eq:4terms_2}. 

To simplify the notation, hereafter, we define  vectors $\underline{\bm h}(t)$, $\underline{\bm g}(t)$, and $\underline{\bm v}(t) $ by
\beq \label{eq:define_3vectors}
\underline{\bm h}(t) \eqdef \left( \begin{array}{c}
 {\bm h}_{\prm}(t) \\
-\sqrt{\frac{\beta}{N}} {\bm h}_{ {\bm w}_1 } (t) \\
\vdots \\
-\sqrt{\frac{\beta}{N}} {\bm h}_{ {\bm w}_N } (t) 
\end{array}
\right),~
\underline{\bm g}(t) \eqdef \left( \begin{array}{c}
 {\bm g}_{\prm}(t) \\
-\sqrt{\frac{\beta}{N}} {\bm g}_{ {\bm w}_1 } (t) \\
\vdots \\
-\sqrt{\frac{\beta}{N}} {\bm g}_{ {\bm w}_N } (t) 
\end{array}
\right),~
\underline{\bm v}(t) \eqdef 
\left( \begin{array}{c}
\bar{\prm}(t) - \prm^\star \\
\frac{1}{\sqrt{\beta N}} \big( {\bm w}_1^t - {\bm w}_1^\star \big) \\
\vdots \\
\frac{1}{\sqrt{\beta N}} \big( {\bm w}_N^t - {\bm w}_N^\star \big)
\end{array}
\right) \eqs.
\eeq
Using \eqref{eq:opt_cond}, it can be verified that (see the detailed  derivation in
Section~\ref{sec:detailed})
\beq \label{eq:new_equality}
\underline{\bm h}(t) = {\bm G} \underline{\bm v}(t) \eqs.
\eeq
By adopting the analysis in \citep{du2017stochastic} and under 
Assumption \ref{ass:fr}, it can be shown that with
\beq\notag
\beta \eqdef \frac{ 8 ( \rho + \lambda_{\sf max} \bigl [  \hat{\bm A}^\top \hat{\bm C}^{-1} \hat{\bm A} )  \bigr ]}{ \lambda_{\sf min}( \hat{\bm C}) } \eqs,
\eeq
then ${\bm G}$ is full rank with its eigenvalues satisfying
\beq\label{eq:eigen_G}
\lambda_{\sf max} ( {\bm G} ) \leq \left| \frac{\lambda_{\sf max} ( \hat{\bm C} ) }{
\lambda_{\sf min} (\hat{\bm C} )} \right| \lambda_{\sf max} ( \rho {\bm I} + \hat{\bm A}^\top \hat{\bm C}^{-1} \hat{\bm A} ),\qquad 
\lambda_{\sf min} ( {\bm G} ) \geq \frac{8}{9} \lambda_{\sf min} ( \hat{\bm A}^\top \hat{\bm C}^{-1} \hat{\bm A} ) > 0 \eqs. 
\eeq 
Moreover,  let 
$
{\bm G} \eqdef {\bm U} \bm{\Lambda} {\bm U}^{-1} \eqs.
$ 
be the eigen-decomposition of ${\bm G}$, where ${\bm \Lambda}$ is  a diagonal matrix consists of the eigenvalues of ${\bm G}$, and the columns of ${\bm U}$ are the eigenvectors. 
Then,  ${\bm U}$ is full rank with
\beq\label{eq:eigen_U}
\| {\bm U} \| \leq 8 \bigl [  \rho + \lambda_{\sf max} ( \hat{\bm A}^\top \hat{\bm C}^{-1} \hat{\bm A} ) \bigr ]\left| \frac{\lambda_{\sf max} ( \hat{\bm C} ) }{
\lambda_{\sf min} (\hat{\bm C} )} \right| ,\qquad \| {\bm U}^{-1} \| \leq \frac{1}{ \rho + \lambda_{\sf max} ( \hat{\bm A}^\top \hat{\bm C}^{-1} \hat{\bm A} )} \eqs.
\eeq
Furthermore, we also define the following upper bounds on the spectral norms 
\beq\label{eq:norm_bounds}
G \eqdef \| {\bm G} \|, \quad \overline{G} \eqdef \max_{p=1,...,M} \| {\bm G}_p \|,\quad \overline{A} \eqdef \max_{p=1,...,M} \| {\bm A}_p \|, \quad \overline{C} \eqdef \max_{p=1,...,M} \| {\bm C}_p \| \eqs.
\eeq
Lastly, we define the following two Lyapunov functions 
\beq \label{eq:lya_funcs}
{\cal E}_c(t) \eqdef \frac{1}{N}  \biggl [ \sum_{i=1}^N \| \prm_i^t - \overline{\prm}(t) \|^2 \biggr ]^{1/2 } ,\qquad {\cal E}_g(t) \eqdef \frac{1}{N} \biggl [ \sum_{i=1}^N  \| {\bm s}_i^t - {\bm g}_{\prm}(t) \|^2 \biggr ] ^{1/2} \eqs.
\eeq
Note that we have the following inequalities:
\beq \label{eq:equiv_norm}
 {\cal E}_c(t) \leq \frac{1}{N} \sum_{i=1}^N \| \prm_i^t - \overline{\prm}(t) \|,~~\frac{1}{N}\sum_{i=1}^N \| \prm_i^t - \overline{\prm}(t) \| \leq \sqrt{N} {\cal E}_c(t) \eqs,
\eeq
which follows from the norm equivalence $\| {\bm x} \|_2 \leq \| {\bm x} \|_1 \leq \sqrt{N} \| {\bm x} \|_2$ for any ${\bm x} \in \RR^N$.

\paragraph{Convergence Analysis}
We denote that $\gamma_1 = \gamma$ and $\gamma_2 = \beta \gamma$. 
To study the linear convergence of the {\sf PD-DistIAG} method, 
our first step is to establish a bound on the difference 
from the primal-dual optimal solution, 
$\underline{\bm v}(t)$. Observe with the choice of our step size ratio,
\beq \label{eq:taking_norm}
\underline{\bm v}(t+1) = ( {\bm I} - \gamma {\bm G} ) \underline{\bm v}(t) + \gamma \big[  \underline{\bm h}(t) - \underline{\bm g}(t) \big]  \eqs.
\eeq
Consider the difference vector
$\underline{\bm h}(t) - \underline{\bm g}(t)$. Its first block can be evaluated as 
\beq \label{eq:last1}
\begin{split}
& \big[ \underline{\bm h}(t) - \underline{\bm g}(t) \big]_1 
 = \frac{1}{NM} \sum_{i=1}^N \sum_{p=1}^M \Big[ \rho \big( \bar{\prm}(t) - \prm_i^{\tau_p^t} \big)
+ {\bm A}_p^\top \big( {\bm w}_i^t - {\bm w}_i^{\tau_p^t} \big) \Big] \\
& = \frac{1}{NM} \sum_{i=1}^N \sum_{p=1}^M \Big[  \rho \big( \bar{\prm}(t) - \bar{\prm} (\tau_p^t) \big)
+ {\bm A}_p^\top \big( {\bm w}_i^t - {\bm w}_i^{\tau_p^t} \big) \Big] 
+ \frac{\rho}{NM} \sum_{i=1}^N \sum_{p=1}^M \big( \bar{\prm}( \tau_p^t) - \prm_i^{\tau_p^t} \big) \eqs.
\end{split}
\eeq
Meanwhile, for any $i \in \{ 1, \ldots, N\}$,  the $(i+1)$-th block  is 
\begin{align} \label{eq:last2}
& \big[ \underline{\bm h}(t) - \underline{\bm g}(t) \big]_{i+1}
 = \sqrt{\frac{\beta}{N}} \frac{1}{M} \sum_{p=1}^M \Big[  {\bm A}_p \big( \prm_i^{\tau_p^t} - \bar{\prm}(t) \big) + {\bm C}_p \big( {\bm w}_i^t - {\bm w}_i^{\tau_p^t} \big) \Big] \\
 & = 
 \sqrt{\frac{\beta}{N}} \frac{1}{M} \sum_{p=1}^M \Big[ {\bm A}_p \big( \bar{\prm} (\tau_p^t) - \bar{\prm}(t) \big) + {\bm C}_p \big( {\bm w}_i^t - {\bm w}_i^{\tau_p^t} \big) \Big] 
 + \sqrt{\frac{\beta}{N}} \frac{1}{M} \sum_{p=1}^M {\bm A}_p \big( \prm_i^{\tau_p^t} - \bar{\prm}( \tau_p^t ) \big) \eqs. \notag
\end{align}
For ease of presentation, we stack up and denote the residual terms (related to consensus error) 
in \eqref{eq:last1} and \eqref{eq:last2} 
as the vector $\underline{\bm{\mathcal{E}}}_c(t)$. 
That is, the first block of $\underline{\bm{\mathcal{E}}}_c(t)$ is $\rho / (  NM) \cdot \sum_{i=1}^N \sum_{p=1}^M \big( \bar{\prm}( \tau_p^t) - \prm_i^{\tau_p^t} \big)$, and the remaining blocks are given by $ 
 \sqrt{ {\beta} / {N}} \cdot  1 / {M} \cdot  \sum_{p=1}^M {\bm A}_p \big( \prm_i^{\tau_p^t} - \bar{\prm}( \tau_p^t ) \big) $, $\forall i \in \{ 1, \ldots, N\}$. 
 Then by the definition of ${\bm G}_p$ in \eqref{eq:matrix_Gp}, 
we obtain the following simplification:
\beq\label{eq:grad_diff_telescope}
\underline{\bm h}(t) - \underline{\bm g}(t) - \underline{\bm{\mathcal{E}}}_c(t)
= \frac{1}{M} \sum_{p=1}^M {\bm G}_p \Big( {\textstyle \sum_{j=\tau_p^t}^{t-1} \underline{\Delta {\bm v}} (j)} \Big),
\eeq
where we have defined
\beq\label{eq:defineDeltav}
\underline{\Delta {\bm v}} (j) \eqdef 
\left( \begin{array}{c}
\bar{\prm}(j+1) - \bar{\prm}(j) \\
\frac{1}{\sqrt{\beta N}} \big( {\bm w}_1^{j+1} - {\bm w}_1^j \big) \\
\vdots \\
\frac{1}{\sqrt{\beta N}} \big( {\bm w}_N^{j+1} - {\bm w}_N^j \big)
\end{array}
\right) \eqs.
\eeq
Clearly, we can express $\Delta \underline{\bm v}(j)$ as 
$
\Delta \underline{\bm v}(j) = \underline{\bm v}(j+1) - \underline{\bm v}(j)  
$ 
with $\underline {\bm v}(t)$ defined in \eqref{eq:define_3vectors}.
Combining \eqref{eq:new_equality} and \eqref{eq:taking_norm}, we can also write $\Delta \underline{\bm v}(j) $  in \eqref{eq:defineDeltav} as 
\beq\label{eq:some_equation1}
\Delta \underline{\bm v}(j) = \gamma \big[  \underline{\bm h}(j) - \underline{\bm g}(j) \big] - \gamma  \underline{\bm h}(j) \eqs.
\eeq
Denoting $\widehat{\underline{\bm v}}(t) \eqdef {\bm U}^{-1} \underline{\bm v}(t)$, multiplying ${\bm U}^{-1}$ on both sides of \eqref{eq:taking_norm} yields
\beq\label{eq:some_equation2}
\widehat{\underline{\bm v}}(t+1) = ( {\bm I} - \gamma \bm{\Lambda} ) \widehat{\underline{\bm v}}(t) + \gamma~ {\bm U}^{-1} \big( \underline{\bm h}(t) - \underline{\bm g}(t) \big) \eqs.
\eeq
Combining \eqref{eq:grad_diff_telescope}, \eqref{eq:some_equation1}, and \eqref{eq:some_equation2}, by triangle inequality, we have
\begin{align} \label{eq:some_equation3}
& \| \widehat{\underline{\bm v}}(t+1) \|  \leq \\ 
& \qquad  \big\| {\bm I} - \gamma \bm{\Lambda} \big\| 
\| \widehat{\underline{\bm v}}(t) \| 
+ \gamma \| {\bm U}^{-1} \| \biggl \{ \|  \underline{\bm{\mathcal{E}}}_c(t) \|
 + \frac{\gamma \overline{G}}{M} \sum_{p=1}^M \sum_{j=\tau_p^t}^{t-1} \big [  \| \underline{\bm h}(j) \| + \| \underline{\bm h}(j) - \underline{\bm g}(j) \| \big ]  \bigg \} \eqs, \notag
\end{align}
where $\overline {G} $ appears in \eqref{eq:norm_bounds} and $\underline{\bm{\mathcal{E}}}_c(t)$ is the residue  term of the consensus.
Furthermore, simplifying the right-hand side of \eqref{eq:some_equation3} yields 
\begin{align} \label{eq:bound_v_norm_upper}
& \| \widehat{\underline{\bm v}}(t+1) \|   \leq \big\| {\bm I} - \gamma \bm{\Lambda} \big\| 
\| \widehat{\underline{\bm v}}(t) \|  
+ \gamma \|{\bm U}^{-1}\| \biggl \{ \|  \underline{\bm{\mathcal{E}}}_c(t) \|
 + \gamma \overline{G} \sum_{j=(t-M)_+}^{t-1} \big[   \| \underline{\bm h}(j) \| + \| \underline{\bm h}(j) - \underline{\bm g}(j) \| \big ]  \biggr  \}   \notag \\ 
 & \qquad  \leq \big\| {\bm I} - \gamma \bm{\Lambda} \big\| 
\| \widehat{\underline{\bm v}}(t) \| 
+ \gamma \| {\bm U}^{-1} \| \Bigg( \|  \underline{\bm{\mathcal{E}}}_c(t) \| + \gamma \overline{G}  \cdot \\
& \qquad \qquad  \sum_{j=(t-M)_+}^{t-1} \bigg \{  \|  \underline{\bm{\mathcal{E}}}_c(j) \| + G \| {\bm U} \| \| \widehat{\underline{\bm v}}(j) \| + 
\overline{G}  \| {\bm U} \|  \sum_{\ell=(j-M)_+}^{j-1}  \bigl [  \| \widehat{\underline{\bm v}}(\ell+1) \| + \| \widehat{\underline{\bm v}}(\ell) \|  \bigr ]  
 \bigg\}  \Bigg)  \eqs. \notag 
\end{align} 
Moreover, using the definition and \eqref{eq:equiv_norm}, we can 
upper bound  $\| \underline{\bm{\mathcal{E}}}_c(t) \|$ by
\beq\label{eq:Ec_bound}
\| \underline{\bm{\mathcal{E}}}_c(t) \| \leq \frac{ 1 }{ M } \sum_{p=1}^M \bigg[ 
\big( \rho + \overline{A} \sqrt{\beta N} \big)  \frac{1}{N} \sum_{i=1}^N
\| \prm_i^{\tau_p^t} - \bar{\prm} ( \tau_p^t ) \|  \bigg] 
\leq 
\sqrt{N} \big( \rho + \overline{A} \sqrt{\beta N} \big) \max_{ (t-M)_+ \leq q \leq t } {\cal E}_c(q)  \eqs .
\eeq
Thus, combining \eqref{eq:bound_v_norm_upper} and  \eqref{eq:Ec_bound}, we bound $\| \widehat{\underline{\bm v}}(t+1)  \| $ by
\beq \label{eq:v_1}
 \| \widehat{\underline{\bm v}}(t+1)  \|  \leq 
\big\| {\bm I} - \gamma \bm{\Lambda} \big\| 
\| \widehat{\underline{\bm v}}(t) \| 
+ C_1( \gamma ) \max_{ (t-2M)_+ \leq q \leq t-1 } \| \widehat{\underline{\bm v}}(q) \|
+ C_2( \gamma ) \max_{ (t-2M)_+ \leq q \leq t } {\cal E}_c(q) \eqs ,
\eeq
where constants $C_1(\gamma)$ and $C_2(\gamma)$ are given by
\beq\notag
C_1( \gamma )  \eqdef \gamma^2~ \| {\bm U} \|   \| {\bm U}^{-1} \|   \overline{G} M  \big( G + 2 \overline{G} M \big),~ C_2( \gamma ) \eqdef \gamma  \| {\bm U}^{-1}  \| \big( 1 + \gamma\overline{G} M \big) \sqrt{N} \big( \rho + \overline{A} \sqrt{\beta N} \big).
\eeq
Notice that since ${\bm U}^{-1}$ is full rank, the squared norm 
$\| \widehat{\underline{\bm v}}(t)  \|^2$ is proportional to 
$\| \overline{\prm}(t) - \prm^\star \|^2 + (1/\beta N) \sum_{i=1}^N \| {\bm w}_i^\star - {\bm w}_i^t \|^2$, \ie
the optimality gap at the $t$-th iteration.

We next  upper bound ${\cal E}_c(t+1)$ as defined in \eqref{eq:lya_funcs}. Notice that $N {\cal E}_c(t+1)$ can be written as Frobenius norm of the matrix $\bm{\Theta}^{t+1} - {\bf 1} \overline{\prm}(t+1)^\top$, where $\bm{\Theta}^{t+1} = ( (\prm_1^{t+1})^\top ; \cdots ; (\prm_N^{t+1})^\top )$. Also, we denote ${\bm S}^t = ( ({\bm s}_1^t)^\top; \cdots ({\bm s}_N^t)^\top)$.  By  the   update in \eqref{eq:pd_alg} and using the triangle inequality, we have
\beq \label{eq:fin_c02}
\begin{split}
{\cal E}_c(t+1) & = \frac{1}{N} \| \bm{\Theta}^{t+1} - {\bf 1} \overline{\prm}(t+1)^\top \|_F
= \frac{1}{N} \| {\bm W} ( \bm{\Theta}^{t} - {\bf 1} \overline{\prm}(t)^\top ) - \gamma ( {\bm S}^t - {\bf 1} {\bm g}_{\prm}(t)^\top ) \|_F \\
& \leq \frac{1}{N} \big( \| \bm{\Theta}^{t+1} - {\bf 1} \overline{\prm}(t)^\top \|_F + \gamma \!~\| {\bm S}^t - {\bf 1} {\bm g}_{\prm}(t)^\top \|_F \big) \eqs.
\end{split}
 \eeq
Notice that we have $\lambda \eqdef \lambda_{\sf max}({\bm W} - (1/N) {\bf 1}{\bf 1}^\top ) < 1$ as the graph is connected. Using the fact that $N {\cal E}_g(t) = \| {\bm S}^t - {\bf 1} {\bm g}_{\prm}(t)^\top  \|_F$, 
the right-hand side of \eqref{eq:fin_c02} can be bounded by 
 \begin{align}\label{eq:fin_c2}
 {\cal E}_c(t+1) & \leq  
\lambda ~ {\cal E}_c(t) + \gamma ~ {\cal E}_g(t) \eqs,
\end{align}
where the Lyapunov function ${\cal E}_g(t) $ is defined in \eqref{eq:lya_funcs}.

To conclude the proof, we need to further  upper bound ${\cal E}_g(t+1)$.   To simplify the notation, let us define ${\bm G}_{p}^t = ( \grd_{\prm} J_{1,p} (\prm_1^t;{\bm w}_1^t)^\top; \cdots ;  \grd_{\prm} J_{N,p} (\prm_N^t;{\bm w}_N^t)^\top )$ and observe that
\beq
{\cal E}_g(t+1) = \frac{1}{N} \Big\| {\bm S}^{t+1} - {\bf 1}{\bm g}_{\prm}(t+1)^\top \Big\|_F
= \frac{1}{N} \Big\| {\bm W} {\bm S}^{t} + {\textstyle \frac{1}{M}} \big( {\bm G}_{p_{t+1}}^{t+1} - {\bm G}_{p_{t+1}}^{\tau_{p_{t+1}}^t} \big) - {\bf 1}{\bm g}_{\prm}(t+1)^\top \Big\|_F
\eeq
where we have used \eqref{eq:s_upd}. 
Furthermore, we observe
\begin{align} \label{eq:bound_Eg_first}
{\cal E}_g(t+1) & = \frac{1}{N} \Big\| {\bm W} ( {\bm S}^{t} -  {\bf 1} {\bm g}_{\prm}(t)^\top )+ {\textstyle \frac{1}{M}} \big( {\bm G}_{p_{t+1}}^{t+1} - {\bm G}_{p_{t+1}}^{\tau_{p_{t+1}}^t} \big) - {\bf 1}( {\bm g}_{\prm}(t+1) - {\bm g}_{\prm}(t) )^\top \Big\|_F \notag  \\
& \leq \frac{1}{N} \Big( \| {\bm W} ( {\bm S}^t - {\bf 1} {\bm g}_{\prm}(t)^\top ) \|_F + \| {\textstyle \frac{1}{M}} \big( {\bm G}_{p_{t+1}}^{t+1} - {\bm G}_{p_{t+1}}^{\tau_{p_{t+1}}^t} \big) - {\bf 1}({\bm g}_{\prm}(t+1) - {\bm g}_{\prm}(t))^\top \|_F \Big) \notag\\
& \leq \lambda \!~ {\cal E}_g(t) + \frac{1}{N} \Bigl \| {\textstyle \frac{1}{M}} \big( {\bm G}_{p_{t+1}}^{t+1} - {\bm G}_{p_{t+1}}^{\tau_{p_{t+1}}^t} \big) - {\bf 1}({\bm g}_{\prm}(t+1) - {\bm g}_{\prm}(t))^\top \Bigr \|_F \eqs . 
\end{align}
We observe 
${\bm G}_{p}^t  = (  ({\bm w}_1^t)^\top {\bm A}_p; \cdots ; ({\bm w}_N^t)^\top {\bm A}_p ) + \rho \!~ \bm{\Theta}^t$ 
and
${\bm g}_{\prm}(t+1) - {\bm g}_{\prm} (t)  = M^{-1} \big( \rho \!~ \bar{\prm}(t+1) - \rho \!~ \bar{\prm}( \tau_{p_{t+1}}^t ) + N^{-1} {\bm A}_{p_{t+1}}^\top \sum_{i=1}^N \big( {\bm w}_i^{t+1} - {\bm w}_i^{\tau_{p_{t+1}}^t} \big) \big)$. Adopting the notations $\bm{\Omega}^t = ( ({\bm w}_1^t)^\top ; \cdots; ({\bm w}_N^t )^\top )$ and $\overline{\bm w}^t = N^{-1} \sum_{i=1}^N {\bm w}_i^t$, we observe that
\beq
\begin{split}
& M^{-1} ( {\bm G}_{p_{t+1}}^{t+1} - {\bm G}_{p_{t+1}}^{\tau_{p_{t+1}}^t} ) - {\bf 1} ( {\bm g}_{\prm}(t+1) - {\bm g}_{\prm}(t) )^\top \\
& \qquad = \frac{\rho}{M} \big( \bm{\Prm}^{t+1} - {\bf 1} \overline{\prm}(t+1)^\top - \bm{\Prm}^{\tau_{p_{t+1}}^t} - {\bf 1} \overline{\prm}(\tau_{p_{t+1}}^t)^\top \big) \\
& \quad \qquad+ \frac{1}{M} \big( 
\bm{\Omega}^{t+1} - {\bf 1} (\overline{\bm w}^{t+1})^\top - \bm{\Omega}^{ \tau_{p_{t+1}}^t } + {\bf 1} (\overline{\bm w}^{\tau_{p_{t+1}}^t})^\top \big) {\bm A}_{p_{t+1}} \eqs .
\end{split}
\eeq
Using the triangular inequality,  the norm of the above can be bounded as
\beq
\begin{split}
& \frac{\rho}{M} \Big( \| \bm{\Prm}^{t+1} - {\bf 1} \overline{\prm}(t+1)^\top \|_F + \| \bm{\Prm}^{\tau_{p_{t+1}}^t} - {\bf 1} \overline{\prm}(\tau_{p_{t+1}}^t)^\top \|_F \Big) \\
& \quad + \frac{ \| {\bm A}_{p_{t+1}} \| }{ M } \Big( \| \bm{\Omega}^{t+1} - {\bf 1} (\overline{\bm w}^{t+1})^\top - \bm{\Omega}^{ \tau_{p_{t+1}}^t } + {\bf 1} (\overline{\bm w}^{\tau_{p_{t+1}}^t})^\top \|_F \Big) \eqs.
\end{split}
\eeq
From the norm equivalence $\| {\bm x} \|_2 \leq \| {\bm x} \|_1$, we recognize that 
$\| \bm{\Omega}^{t+1} - {\bf 1} (\overline{\bm w}^{t+1})^\top - \bm{\Omega}^{ \tau_{p_{t+1}}^t } + {\bf 1} (\overline{\bm w}^{\tau_{p_{t+1}}^t})^\top \|_F
\leq \sum_{i=1}^N \| {\bm w}_i^{t+1} - \overline{\bm w}^{t+1} 
- {\bm w}_i^{\tau_{p_{t+1}}^t } + \overline{\bm w}^{\tau_{p_{t+1}}^t } \|$. 
It holds for all 
 $t' \leq t$  that
\beq \notag
{\bm w}_i^{t+1} - {\bm w}_i^{t'} = - \frac{\gamma}{ \beta M} \sum_{\ell=t'}^t \sum_{p=1}^M \Big[ {\bm A}_p ( \prm_i^{\tau_p^\ell} - \prm^\star ) - {\bm C}_p ( {\bm w}_i^{\tau_p^\ell} - {\bm w}_i^\star ) \Big]  \eqs .
\eeq
We thus obtain
\begin{align}\label{eq:bound_sum_delta_tt}
& \frac{ \| {\bm A}_{p_{t+1}} \| }{ M } \Big( \| \bm{\Omega}^{t+1} - {\bf 1} (\overline{\bm w}^{t+1})^\top - \bm{\Omega}^{ \tau_{p_{t+1}}^t } + {\bf 1} (\overline{\bm w}^{\tau_{p_{t+1}}^t})^\top \|_F \Big) \notag \\
& \qquad \leq \frac{ \| {\bm A}_{p_{t+1}} \| }{ M } \sum_{i=1}^N
\bigg\| {\bm w}_i^{t+1} - \frac{1}{N} \sum_{j=1}^N {\bm w}_j^{t+1} 
- {\bm w}_i^{\tau_{p_{t+1}}^t } + \frac{1}{N} \sum_{j=1}^N {\bm w}_j^{\tau_{p_{t+1}}^t } \bigg\| 
\notag \\
&\qquad  \leq 
\frac{ 2 \gamma \overline{A} }{ \beta M^2} \sum_{i=1}^N \sum_{\ell = (t-M)_+}^t  \sum_{p=1}^M \Big(
\big \| {\bm A}_p ( \prm_i^{\tau_p^\ell} - \prm^\star ) - {\bm C}_p ( {\bm w}_i^{\tau_p^\ell} - {\bm w}_i^\star ) \big\| \Big)  \notag \\
& \qquad \leq \frac{ 2\gamma \overline{A} }{ \beta M} \sum_{i=1}^N \sum_{\ell = (t-M)_+}^t  \biggl [ \max_{ (\ell-M)_+ \leq q \leq \ell } \Big( \overline{A} \| \prm_i^q - \prm^\star \| + \overline{C} \| {\bm w}_i^q - {\bm w}_i^\star \| \Big) \biggr ] \eqs.
\end{align}
Thus, combining \eqref{eq:fin_c2}, 
\eqref{eq:bound_sum_delta_tt}, and the definition of ${\cal E}_c$ in \eqref{eq:lya_funcs}, we have 
\begin{align}  \label{eq:bound_delta_avg}
 & \textstyle  \frac{1}{N} \| {\textstyle \frac{1}{M}} \big( {\bm G}_{p_{t+1}}^{t+1} - {\bm G}_{p_{t+1}}^{\tau_{p_{t+1}}^t} \big) - {\bf 1}({\bm g}_{\prm}(t+1) - {\bm g}_{\prm}(t))^\top \|_F \\
 & \notag \leq \frac{\rho}{M} \bigl [  {\cal E}_c ( \tau_{p_{t+1}}^t ) + {\cal E}_c(t+1)  \bigr ] \\
 & \qquad \qquad  +  \frac{ 2 \gamma \overline{A} (M+1) }{ \beta NM} \sum_{i=1}^N
\max_{ (t-2M)_+ \leq q \leq t } \Big( \overline{A} \| \prm_i^q - \prm^\star \| + \overline{C} \| {\bm w}_i^q - {\bm w}_i^\star \| \Big)  \notag \\
&   \leq   \frac{ \rho }{M} \bigl [  {\cal E}_c ( \tau_{p_{t+1}}^t )  
+ \lambda  \!~ {\cal E}_c ( t ) + \gamma~ {\cal E}_g ( t ) \bigr ] \notag \\
& \notag \qquad   + \frac{ 2 \gamma \overline{A} (M+1) }{ \beta M}
\max_{ (t-2M)_+ \leq q \leq t } \Big( \overline{A}~\sqrt{N} {\cal E}_c(q) + \overline{A}~\| \bar{\prm}(q) - \prm^\star\|  + \frac{ \overline{C} }{N} \sum_{i=1}^N \| {\bm w}_i^q - {\bm w}_i^\star \| \Big) \eqs.
\end{align}
Combining \eqref{eq:bound_Eg_first} and \eqref{eq:bound_delta_avg}, we obtain that \beq
\begin{split}\label{eq:final_bound_Eg2}
& {\cal E}_g(t+1)  \leq \Big( \lambda + \frac{ \gamma \rho }{M} \Big) ~{\cal E}_g( t) + \bigg[ 
\frac{ 2 \gamma \overline{A}^2 (M+1) \sqrt{N} }{\beta M} + \frac{2 (1 + \lambda)}{M} \bigg]
\max_{ (t-2M)_+ \leq q \leq t } {\cal E}_c(q) \\
& \hspace{2.25cm} + \frac{ 2 \gamma \overline{A} (M+1) }{\beta M}
\max_{ (t-2M)_+ \leq q \leq t } \bigg( \overline{A}~\| \bar{\prm}(q) - \prm^\star\|  + \frac{ \overline{C} }{N} \sum_{i=1}^N \| {\bm w}_i^q - {\bm w}_i^\star \| \bigg) \eqs.
\end{split}
\eeq
To bound the last term on the right-hand side of \eqref{eq:final_bound_Eg2}, 
For all $q$, we observe that:
\beq \notag 
\begin{split}
& \bigg( \overline{A}~\| \bar{\prm}(q) - \prm^\star\|  + \frac{ \overline{C} }{N} \sum_{i=1}^N \| {\bm w}_i^q - {\bm w}_i^\star \| \bigg)^2 \\
&  \qquad \leq (N+1) (\overline{A})^2 \bigg[  \| \bar{\prm}(q) - \prm^\star\|^2 + \beta \Big( \frac{\overline{C}}{\overline{A}} \Big)^2 \frac{1}{ \beta N} \sum_{i=1}^N \| {\bm w}_i^q - {\bm w}_i^\star \|^2 \bigg]\\
& \qquad \leq (N+1) ~\| {\bm U} \| \max \Big\{ (\overline{A})^2, \beta (\overline{C})^2 \Big\}~  \| \underline{\bm v}(q) \|^2 \eqs, 
\end{split}
\eeq
which further implies that 
\beq\label{eq:eg_fin}
\begin{split}
& {\cal E}_g(t+1) \leq \Big( \lambda + \frac{ \gamma \rho }{M} \Big) ~{\cal E}_g( t) + \Big(
\frac{ 2 \gamma \overline{A}^2 (M+1) \sqrt{N} }{\beta M} + \frac{2 (1 + \lambda)}{M} \Big) 
\max_{ (t-2M)_+ \leq q \leq t } {\cal E}_c(q) \\
& \hspace{3.15cm} + \frac{ 2 \gamma \overline{A} \sqrt{N+1} (M+1) }{\beta M} ~\| {\bm U} \|
\max\{ \overline{A}, \sqrt{\beta} \overline{C} \}
\max_{ (t-2M)_+ \leq q \leq t } \| \widehat{\underline{\bm v}}(q) \| \eqs.
\end{split}
\eeq
Finally, combining \eqref{eq:v_1}, \eqref{eq:fin_c2}, \eqref{eq:eg_fin} shows:
\beq \label{eq:fin_sys}
\left( 
\begin{array}{c}
\| \widehat{\underline{\bm v}}(t+1) \| \\[.1cm]
{\cal E}_c(t+1) \\[.1cm]
{\cal E}_g(t+1) 
\end{array}
\right)
\leq 
{\bm Q} 
\left( 
\begin{array}{c}
\max_{ (t-2M)_+ \leq q \leq t }  \| \widehat{\underline{\bm v}}(q) \| \\[.1cm]
\max_{ (t-2M)_+ \leq q \leq t }  {\cal E}_c(q) \\[.1cm]
\max_{ (t-2M)_+ \leq q \leq t }  {\cal E}_g(q) 
\end{array}
\right) \eqs,
\eeq 
where
the inequality sign is applied element-wisely,
and ${\bm Q}$ is a non-negative $3 \times 3$ matrix, defined by:
\beq \label{eq:defq}
\left( 
\begin{array}{ccc}
	\theta (\gamma) + \gamma^2 \| {\bm U} \| \| {\bm U}^{-1} \| \overline{G} M (  G + 2 \overline{G} M ) & \gamma \sqrt{N} \| {\bm U} \| ( 1 + \gamma \overline{G} M ) ( \rho + \overline{A} \sqrt{\beta N} ) & 0 \\[.1cm]
	0 & \lambda & \gamma \\[.1cm]
	\frac{ 2 \gamma \overline{A} \sqrt{N+1} (M+1) }{\beta M} ~\| {\bm U} \|
\max\{ \overline{A}, \sqrt{\beta}~ \overline{C} \}
 & 
	\sqrt{N} \frac{ 2 \gamma \overline{A}^2 (M+1) }{\beta M} + \frac{2 (1 + \lambda)}{M} & 
	\lambda + \frac{\gamma \rho}{M} 
\end{array}
\right)\eqs,
\eeq
where $  \theta( \gamma ) \eqdef \| {\bm I} - \gamma {\bm \Lambda } \| = \| {\bm I} - \gamma {\bm G} \| $. Note that the upper bounds for  $ \| {\bm U} \|$ and  $  \| {\bm U}^{-1} \|$ are provided in \eqref{eq:eigen_U}. 
Furthermore, also note that  the eigenvalues of $\bm G$ are bounded in \eqref{eq:eigen_G}. We could set the stepsize $\gamma$ to be sufficiently small such that 
such that $\theta( \gamma ) \eqdef \| {\bm I} - \gamma {\bm G} \| < 1$.  

Finally, we apply  Lemmas ~\ref{lem:ineq} and \ref{lem:exist} presented 
in Section~\ref{sec:useful} to the recursive inequality in  \eqref{eq:eg_fin}, which 
shows that 
each of $\| \underline{\bm v}(t) \|$, ${\cal E}_c(t)$, ${\cal E}_g(t)$ 
converges linearly with $t$. Therefore, we conclude the proof of  Theorem~\ref{thm:main}.


\subsection{Two Useful Lemmas} \label{sec:useful}
In this section, we present two auxiliary lemmas that are used in the proof of Theorem~\ref{thm:main}.  Our first lemma establish the linear convergence of vectors satisfying recursive relations similar to  \eqref{eq:eg_fin}, provided the spectral radius of ${\bm Q}$ is less than one. In addition, the second lemma verifies this condition for ${\bm Q}$ defined in  in \eqref{eq:defq}.
\begin{Lemma} \label{lem:ineq}
Consider a sequence of non-negative vectors $\{ {\bm e}(t) \}_{t \geq 1} \subseteq \RR^n $ whose evolution is characterized by ${\bm e}(t+1) 
\leq 
{\bm Q}\!~
{\bm e}( [ (t-M+1)_+ , t ] )$ 
 for all $t \geq 1$ and some fixed integer $M>0$, where ${\bm Q} \in \RR^{n \times n}$ is a matrix whose entries are nonnegative, and we define 
\beq \notag 
{\bm e}( {\cal S} ) \eqdef
\left( \begin{array}{c}
\max_{ q \in {\cal S} } ~e_1 (q) \\
\vdots \\
\max_{ q \in {\cal S} } ~e_n(q) 
\end{array} \right) \in \RR^n \qquad \text{for any  subset}~{\cal S} \subseteq \NN\eqs .
\eeq
Moreover,  
if  ${\bm Q} $  irreducible in the sense that there exists an integer $m$ such that the entries of ${\bm Q}^m$ are all positive,  and the spectral radius of ${\bm Q}$, denoted by $\rho(  {\bm Q} )$,  is strictly  less than one, then for any $t\geq 1$, we have
\beq \label{eq:q_result}
{\bm e}(t) \leq \rho({\bm Q})^{\lceil \frac{t-1}{ M} \rceil} C_1 {\bm u}_1 \eqs,
\eeq
where ${\bm u}_1 \in \RR_{++}^n$ is the top right eigenvector of ${\bm Q}$ 
and $C_1$ is a constant that depends on the initialization.
\end{Lemma}

{\bf Proof}. 
We shall prove the lemma using induction. By the Perron-Frobenius 
theorem, the eigenvector ${\bm u}_1$ associated with $\rho({\bm Q})$
is unique and is an all-positive vector. Therefore, there exists $C_1$ such that
\beq\label{eq:case00}
{\bm e}(1) \leq C_1\!~ {\bm u}_1 \eqs.
\eeq

Let us first consider the base case with $t=2,...,M+1$, i.e., $\lceil (t-1) / { M} \rceil = 1$.
When $t=2$, by \eqref{eq:case00} we have,
\beq\label{eq:case01}
{\bm e}(2) \leq {\bm Q} {\bm e}(1) \leq C_1\!~ {\bm Q}\!~ {\bm u}_1 = \rho( {\bm Q})\!~ C_1\!~ {\bm u}_1 \eqs,
\eeq
which is valid as ${\bm Q}$, ${\bm e}(1)$, ${\bm u}_1$ are all non-negative.
Furthermore, we observe that ${\bm e}(2) \leq C_1 {\bm u}_1$. 
Next when $t = 3$, we have 
\beq \notag
{\bm e}(3) \leq {\bm Q} {\bm e}([1,2]) \overset{(a)}{\leq} C_1\!~ {\bm Q}\!~ {\bm u}_1 = \rho( {\bm Q})\!~ C_1\!~ {\bm u}_1 \eqs,
\eeq
where (a) is due to the non-negativity of vectors/matrix and 
the fact ${\bm e}(1), {\bm e}(2) \leq C_1 {\bm u}_1$
as shown in \eqref{eq:case01}. 
Telescoping using similar steps, one can   show   
${\bm e}(t) \leq \rho( {\bm Q} )\!~ C_1\!~ {\bm u}_1$
for any $t=2,...,M+1$. 

For the induction step, let us assume that \eqref{eq:q_result}
holds true for any $t$ up to $t= pM + 1$. That is, we assume that the result holds for all $t$ such that $\lceil (t-1) / { M} \rceil \leq p$.  We shall show that it
also holds for any $t = pM+2, ..., (p+1)M + 1$, i.e., $\lceil (t-1) / { M} \rceil = p+ 1$. 
Observe that
\beq \label{eq:induction_case00}
{\bm e}(pM+2)  \leq {\bm Q} \!~ {\bm e}( [ (p-1)M + 2, pM+1 ] )
\leq C_1 \!~ \rho( {\bm Q})^p {\bm Q} {\bm u}_1  =  \rho( {\bm Q})^{p+1}\!~ C_1 \!~ {\bm u}_1 \eqs,
\eeq
where we have used the induction hypothesis.  It is clear that \eqref{eq:induction_case00} is equivalent to \eqref{eq:q_result} with $t = pM+2$.
Similar upper bound can be obtained for ${\bm e}(pM+3)$ as well. 
Repeating the same steps, 
we show that \eqref{eq:q_result} is true for 
any $t = pM+2, ..., (p+1)M + 1$. 
Therefore, we conclude the proof of this lemma. 
\hfill \textbf{Q.E.D.}\vspace{.2cm}

The following Lemma shows that ${\bm Q}$ defined  in \eqref{eq:defq} satisfies the conditions required in the previous lemma.  Combining these two lemmas yields the final step of the proof of Theorem \ref{thm:main}.
\begin{Lemma} \label{lem:exist}
Consider the matrix ${\bm Q}$ defined in \eqref{eq:defq}, it can be shown that (a) ${\bm Q}$ is an irreducible matrix in $\RR^{3 \times 3} $; (b) there exists a sufficiently
small $\gamma$ such that $\rho( {\bm Q}) < 1$; and (c) as $N,M \gg 1$ and
the graph is geometric, we can set $\gamma = {\cal O}( 1 / \max\{N^2,M^2\})$
and $\rho( {\bm Q}) \leq 1 - {\cal O}( 1 / \max\{N^2,M^2\})$.
\end{Lemma}

{\bf Proof.} Our proof is divided into three parts. The first part shows 
the straightforward irreducibility of ${\bm Q}$; the second part 
gives an upper bound to the spectral radius of ${\bm Q}$; and
the last part derives an asymptotic bound on $\rho ( {\bm Q})$ when
$N,M \gg 1$. 

\paragraph{Irreducibility of ${\bm Q}$}To see that ${\bm Q}$ is irreducible, notice that ${\bm Q}^2 $ is a positive matrix, which could be verified by direct computation. 

\paragraph{Spectral Radius of ${\bm Q}$}
In the sequel, we compute an upper bound 
to the spectral radius of ${\bm Q}$, and show that if $\gamma$
is sufficiently small, then its spectral radius will be strictly less than one.
First we note that $\theta(\gamma) = 1 - \gamma \alpha$ for some $\alpha > 0$ and the network connectivity satisfies $\lambda<1$. 
Also note that $\rho>0$. 
For notational simplicity let us define the following
\begin{align*} 
& a_1  = \| {\bm U} \| \| {\bm U}^{-1} \| \overline{G} M (  G + 2 \overline{G} M ), ~  
a_2 =  \|{\bm U}\| \sqrt{N} ( \rho + \overline{A} \sqrt{\beta N} ), ~ a_3 = \overline{G} M \|{\bm U}\| \sqrt{N} ( \rho + \overline{A} \sqrt{\beta N} ) \nonumber\\
& a_4 = \frac{ 2 \overline{A} \sqrt{N+1} (M+1) }{\beta M}~\| {\bm U} \| 
\max\{ \overline{A}, \sqrt{\beta} \overline{C} \}
, \quad   a_5 =	\frac{ 2 \overline{A}^2 (M+1) \sqrt{N} }{\beta M}, \quad a_6 = \frac{2(1+ \lambda)}{M}.
\end{align*}
With the above shorthand definitions, 
the characteristic polynomial for ${\bm Q}$, denoted by $g\colon \RR\rightarrow \RR$,  is given by
\begin{align*}  
g(\sigma)& = \mbox{det}\left( 
\begin{array}{ccc} 
\sigma - (1- \gamma \alpha + \gamma^2 a_1) & -\gamma a_2 - \gamma^2 a_3 & 0 \\[.1cm]
0 & \sigma-\lambda & -\gamma \\[.1cm]
- \gamma a_4 & 
-\gamma a_5 - a_6 & 
\sigma - \Big( \lambda + \frac{\gamma \rho}{M} \Big)
\end{array}
\right).
\end{align*}
By direct computation, we have
\beq
\begin{split}
g(\sigma) & = (\sigma - (1-\gamma \alpha + \gamma^2 a_1 ) ) \!~ g_0(\sigma) - \gamma^3 ( a_2 + \gamma a_3 ) a_4
\end{split}
\eeq
where
\beq
g_0(\sigma) \eqdef (\sigma - \lambda)^2 - \frac{\gamma \rho}{M} (\sigma-\lambda) - \gamma( \gamma a_5 + a_6) \eqs.
\eeq
Notice that the two roots of the above polynomial can be upper bounded by:
\beq
\lambda + \frac{\gamma \rho}{2M} \pm \biggl[ \bigg( \frac{\gamma \rho \sqrt{N}}{2M} \bigg)^2  + \gamma ( \gamma a_5 + a_6 ) \biggr ] ^{1/2} \leq \overline{\sigma} \eqdef \lambda + \frac{\gamma \rho}{M} + \sqrt{ \gamma( \gamma a_5 + a_6 ) }
\eeq
In particular, for all $\sigma \geq \overline{\sigma}$, we have
\beq
g_0( \sigma) \geq ( \sigma - \overline{\sigma}  )^2 \eqs.
\eeq
Now, let us define
\beq \label{eq:sig_upperbd}
\sigma^\star \eqdef \max \left\{ \frac{\gamma \alpha}{4} + 1 - \gamma \alpha + \gamma^2 a_1, \overline{\sigma} + \gamma \sqrt{\frac{4 (a_2 + \gamma a_3) a_4}{\alpha}} \right\} 
\eeq
Observe that for all $\sigma \geq \sigma^\star$, it holds that
\beq
\begin{split}
g(\sigma) & \geq (\sigma - (1- \gamma \alpha + \gamma^2 a_1) ) ( \sigma - \overline{\sigma} )^2 - \gamma^3 ( a_2 + \gamma a_3 ) a_4 \\
& \geq \frac{\gamma \alpha}{4} \!~ \gamma^2 \frac{4 ( a_2 + \gamma a_3) a_4}{\alpha} - \gamma^3 ( a_2 + \gamma a_3 ) a_4 = 0\eqs.
\end{split}
\eeq
Lastly, observe that $g(\sigma)$ is strictly increasing for all
$\sigma \geq \sigma^\star$. Combining
with the Perron Frobenius theorem shows that 
$\rho( {\bm Q} ) \leq \sigma^\star$. 
Moreover, as $\lambda < 1$ and $\alpha > 0$, there
exists a sufficiently small $\gamma$ such that $\sigma^\star < 1$. 
We conclude that $\rho ( {\bm Q} ) < 1$ in the latter case.

\paragraph{Asymptotic Rate when $M , N \gg 1$} 
We evaluate a sufficient condition on $\gamma$ for the 
proposed algorithm to converge, \ie when $\sigma^\star < 1$. 
Let us consider \eqref{eq:sig_upperbd} and the
first operand in the $\max\{ \cdot\}$. The first operand is guaranteed to 
be less than one if:
\beq \label{eq:sig_1}
\gamma \leq \frac{\alpha}{2 a_1} \Longrightarrow 
\frac{\gamma \alpha}{4} + 1 - \gamma \alpha + \gamma^2 a_1 \leq 1 - \frac{\gamma \alpha}{4} \eqs.
\eeq
Moreover, from the definition of $a_1$, we note that this requires $\gamma = {\cal O}(1/M^2)$ if $M \gg 1$. 

Next, we notice that for geometric
graphs, we have $\lambda = 1 - c/N$ for some positive $c$. Substituting this 
into the second operand in
\eqref{eq:sig_upperbd} gives
\beq \label{eq:sig_2}
1 - \frac{c}{N} + \frac{\gamma \rho}{M} + \sqrt{ \gamma( \gamma a_5 + a_6 ) }
+  \gamma \sqrt{\frac{4 (a_2 + \gamma a_3) a_4}{\alpha}} < 1 \eqs.
\eeq
Therefore, \eqref{eq:sig_1} and \eqref{eq:sig_2} together give a sufficient 
condition
for $\sigma^\star < 1$.

To obtain an asymptotic rate
when $M , N \gg 1$. 
Observe that $a_2 = \Theta( {N})$, $a_3 = \Theta(  {N} M)$, $a_4 = \Theta(\sqrt{N})$, 
$a_5 = \Theta(\sqrt{N})$, $a_6 = \Theta(1/M)$.
Moreover, the condition \eqref{eq:sig_1} gives $\gamma = {\cal O}(1/M^2)$, therefore the left hand side of Eq.~\eqref{eq:sig_2} can be approximated by
\beq
1 - \frac{c}{N} + 
\gamma \!~ \Theta \left( {N}^{\frac{3}{4}} \right) 
+ \sqrt{\gamma} \!~ \Theta ( 1 / \sqrt{M} ) \eqs .
\eeq
Setting the above to $1 - c/(2N)$ requires one to have $\gamma = {\cal O}(1/N^{2})$.

Finally, the above discussions show that 
setting $\gamma = {\cal O}( 1 / \max\{N^{2},M^2\})$ guarantees
that $\sigma^\star < 1$. In particular, we have 
$\sigma^\star \leq \max\{ 1 - \gamma \frac{\alpha}{4}, 1 - c/(2N) \} = 1 - {\cal O}( 1 / \max\{ N^{2}, M^2 \})$.
\hfill \textbf{Q.E.D.}

\subsection{Derivation of  Equation (\ref{eq:new_equality})} \label{sec:detailed}
We  we establish  \eqref{eq:new_equality} with details. Recall that $\underline{\bm h}(t) $ and $\underline{\bm v}(t) $ are defined in \eqref{eq:define_3vectors}. We verify this equation for each block of  $\underline{\bm h}(t) $. 
To begin with, for the first block, for ${\bm h}_{\prm} (t)$ defined in \eqref{eq:4terms_1}, we have 
\beq \notag 
{\bm h}_{\prm} (t) = \rho \bar{\prm}(t) + \frac{1}{N} \sum_{i=1}^N \hat{\bm A}^\top {\bm w}_i^t
 = \rho \big( \bar{\prm}(t) - \prm^\star + \prm^\star \big) + \frac{1}{N} \sum_{i=1}^N \hat{\bm A}^\top {\bm w}_i^t \eqs.
\eeq
Recall from \eqref{eq:opt_cond} that
$\rho \prm^\star = - \frac{1}{N} \sum_{i=1}^N \hat{\bm A}^\top {\bm w}_i^\star$,
which implies that 
\beq \label{eq:fin_eq_1}
{\bm h}_{\prm} (t)
= \rho \big( \bar{\prm}(t) - \prm^\star \big) + \sum_{i=1}^N \sqrt{\frac{\beta}{N}} \hat{\bm A}^\top \!~ \frac{1}{\sqrt{\beta N}} \big( {\bm w}_i^t - {\bm w}_i^\star \big)
= [  {\bm G} \underline{\bm v}(t)  ] _{1}  \eqs,
\eeq
where $ [  {\bm G} \underline{\bm v}(t)  ] _{1} $ denotes the first block of $    {\bm G} \underline{\bm v}(t)  $.

It remains to establish the equation for the remaining blocks. For any $i \in \{ 1, \ldots, N\}$, let us focus on the $i+1$-th block. By the definition of ${\bm h}_{\bm w_i}(t)$ in \eqref{eq:4terms_2}, we have 
\beq\notag
- \sqrt{\frac{\beta}{N}}  {\bm h}_{\bm w_i}(t) = 
- \sqrt{\frac{\beta}{N}} \big( \hat{\bm A} \bar{\prm}(t) - \hat{\bm C} {\bm w}_i^t - \hat{\bm b}_i \big)
= - \sqrt{\frac{\beta}{N}} \big( \hat{\bm A} ( \bar{\prm}(t) - \prm^\star ) + \hat{\bm A} \prm^\star - \hat{\bm C} {\bm w}_i^t - \hat{\bm b}_i \big) \eqs.
\eeq
Again from \eqref{eq:opt_cond}, it holds  that $\hat{\bm A} \prm^\star
= {\bm b}_i + \hat{\bm C} {\bm w}_i^\star$. Therefore,
\beq \label{eq:fin_eq_i}
- \sqrt{\frac{\beta}{N}} \big( \hat{\bm A} \bar{\prm}(t) - \hat{\bm C} {\bm w}_i^t - \hat{\bm b}_i \big) 
= - \sqrt{\frac{\beta}{N}} \hat{\bm A} ( \bar{\prm}(t) - \prm^\star )
+ \beta \hat{\bm C} \!~ \frac{{\bm w}_i^t - {\bm w}_i^\star}{\sqrt{\beta N}} 
=[  {\bm G} \underline{\bm v}(t)  ]_{i+1}\eqs,
\eeq
where $[  {\bm G} \underline{\bm v}(t)  ]_{i+1}$ denotes the $i+1$-th  block of $ {\bm G} \underline{\bm v}(t)$.
Combining \eqref{eq:fin_eq_1} and \eqref{eq:fin_eq_i} gives 
the desired equality. 

\section{Additional Experiments}
An interesting observation from Theorem~\ref{thm:main}
is that the convergence rate of {\sf PD-DistIAG} 
depends on $M$ and the topology of the graph. 
The following experiments will demonstrate the effects of these on the algorithm,
along with the effects of regularization parameter $\rho$. 
\begin{figure}[H]
\centering
\includegraphics[width=.425\linewidth]{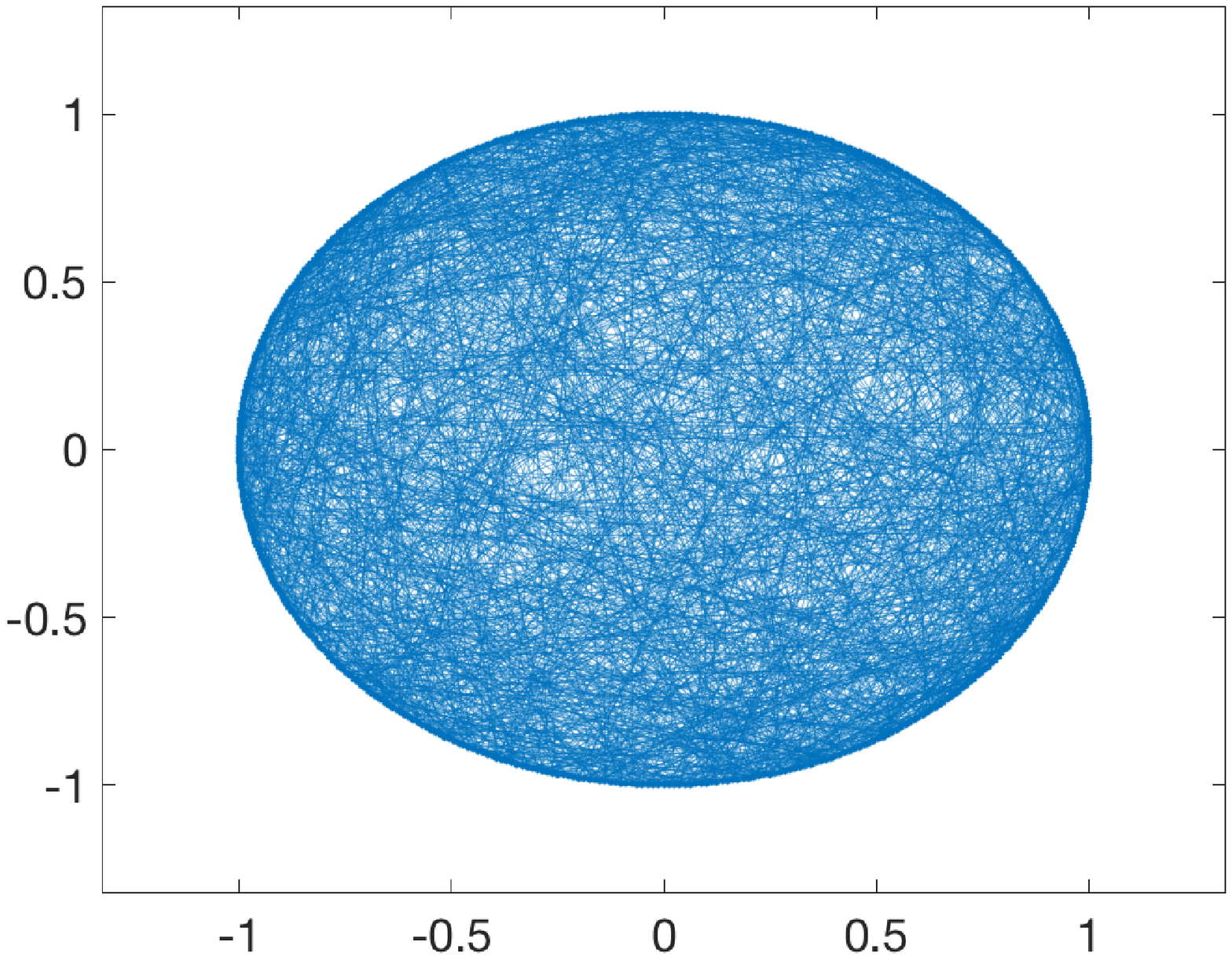}~
\includegraphics[width=.425\linewidth]{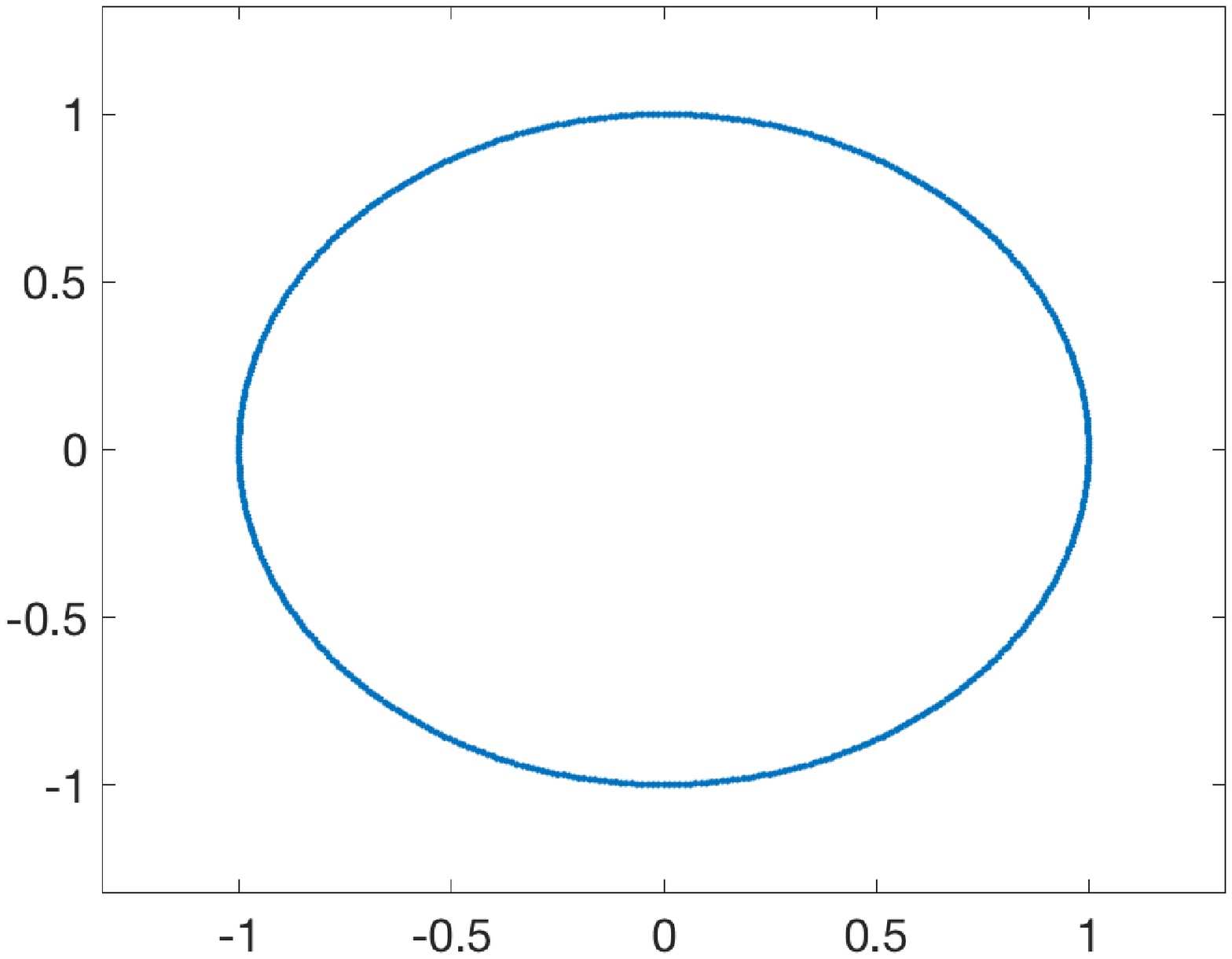}
\caption{Illustrating the graph topologies in the additional experiments. (Left) ER graph with connectivity probability of $1.01\log N / N$. (Right) Ring graph.} \label{fig:graph2}
\end{figure}

\begin{figure}[H]
\centering
\includegraphics[width=.425\linewidth]{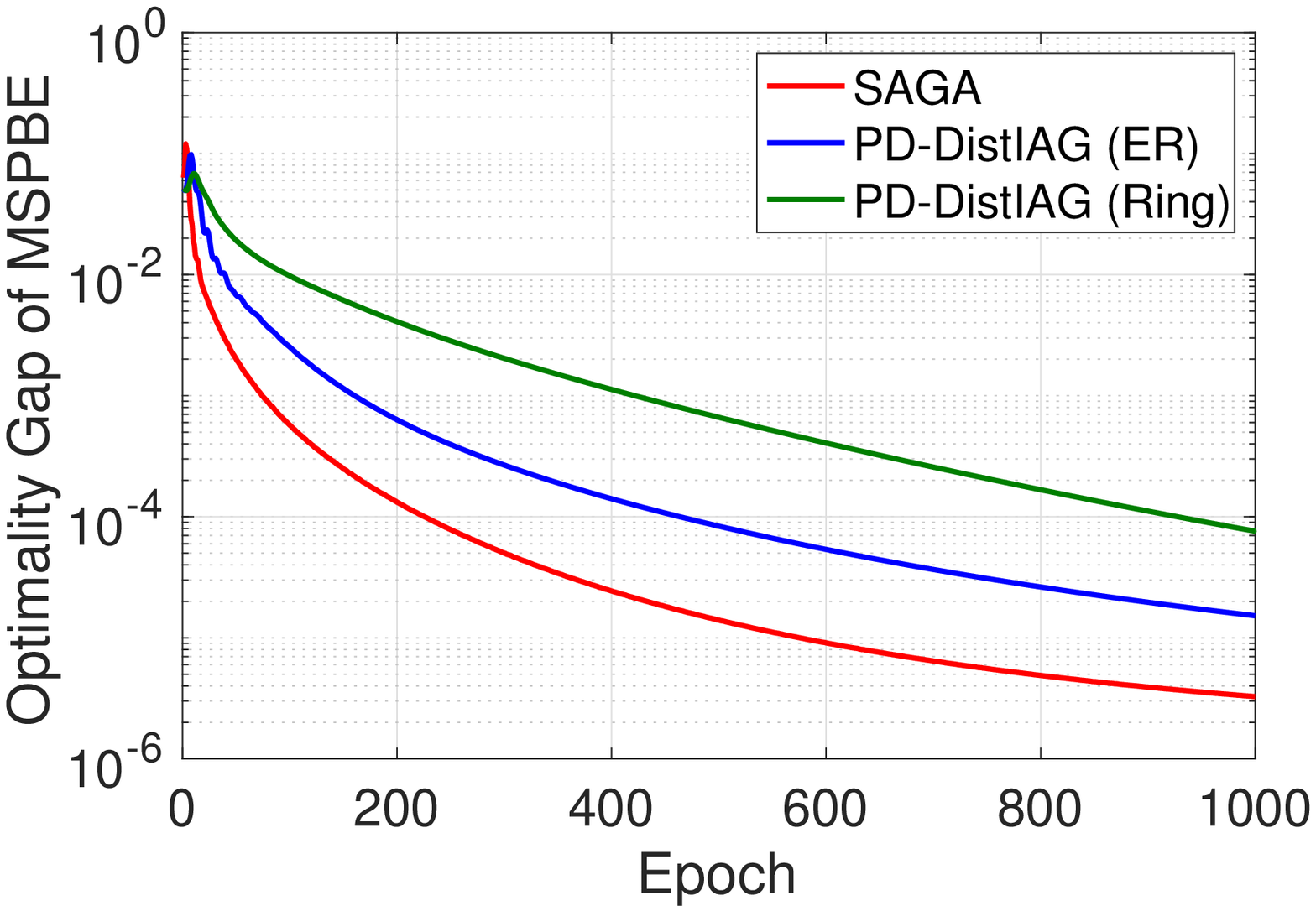}~
\includegraphics[width=.425\linewidth]{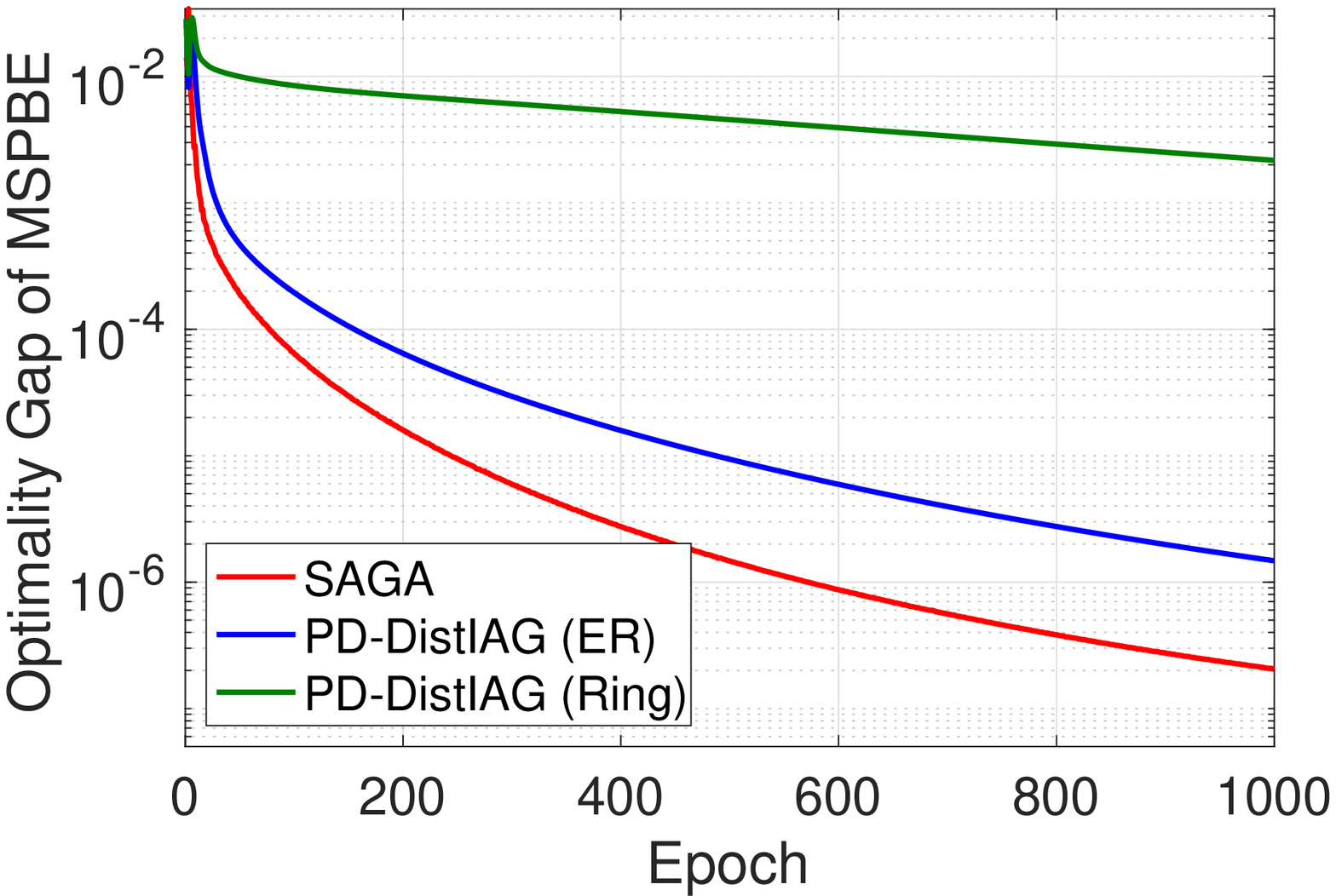}
\caption{Experiment with \texttt{mountaincar} dataset. For this problem, we 
only have  $d=300$, $M=500$ samples, but yet there are $N=500$ agents. 
(Left) We set $\rho = 0.01$. (Right) We set $\rho = 0.1$.} \label{fig:two}
\end{figure}

To demonstrate the dependence of {\sf PD-DistIAG} on the graph topology, 
we fix the number of agents at $N=500$ and compare the performances
on the ring and the
ER graph set with probability of connection of $p= 1.01 \log N / N$, as illustrated
in Fig.~\ref{fig:graph2}.
Notice that the ring graph is not a geometric graph and its connectivity parameter,
defined as $\lambda \eqdef \lambda_{\sf max} ( {\bm W} - (1/N) {\bf 1}{\bf 1}^\top )$ 
from the previous section can be much closer to $1$ than the ER graph. 
Therefore, we expect the {\sf PD-DistIAG} algorithm to converge slower 
on the ring graph. This is corroborated by Fig.~\ref{fig:two}. 
Furthermore, from the figure, we observe that with a larger regularization
$\rho$, the disadvantage for using the ring graph has exacerbated. 
We suspect that this is due to the fact that the convergence speed is limited
by the graph connectivity, as seen in \eqref{eq:defq}; 
while in the case of ER graph, the algorithm
is able to exploit the improved problem's condition number. 

Next, we consider the same set of experiment but increase the number of samples
to $M=5000$. 
\begin{figure}[htpb]
\centering
\includegraphics[width=.425\linewidth]{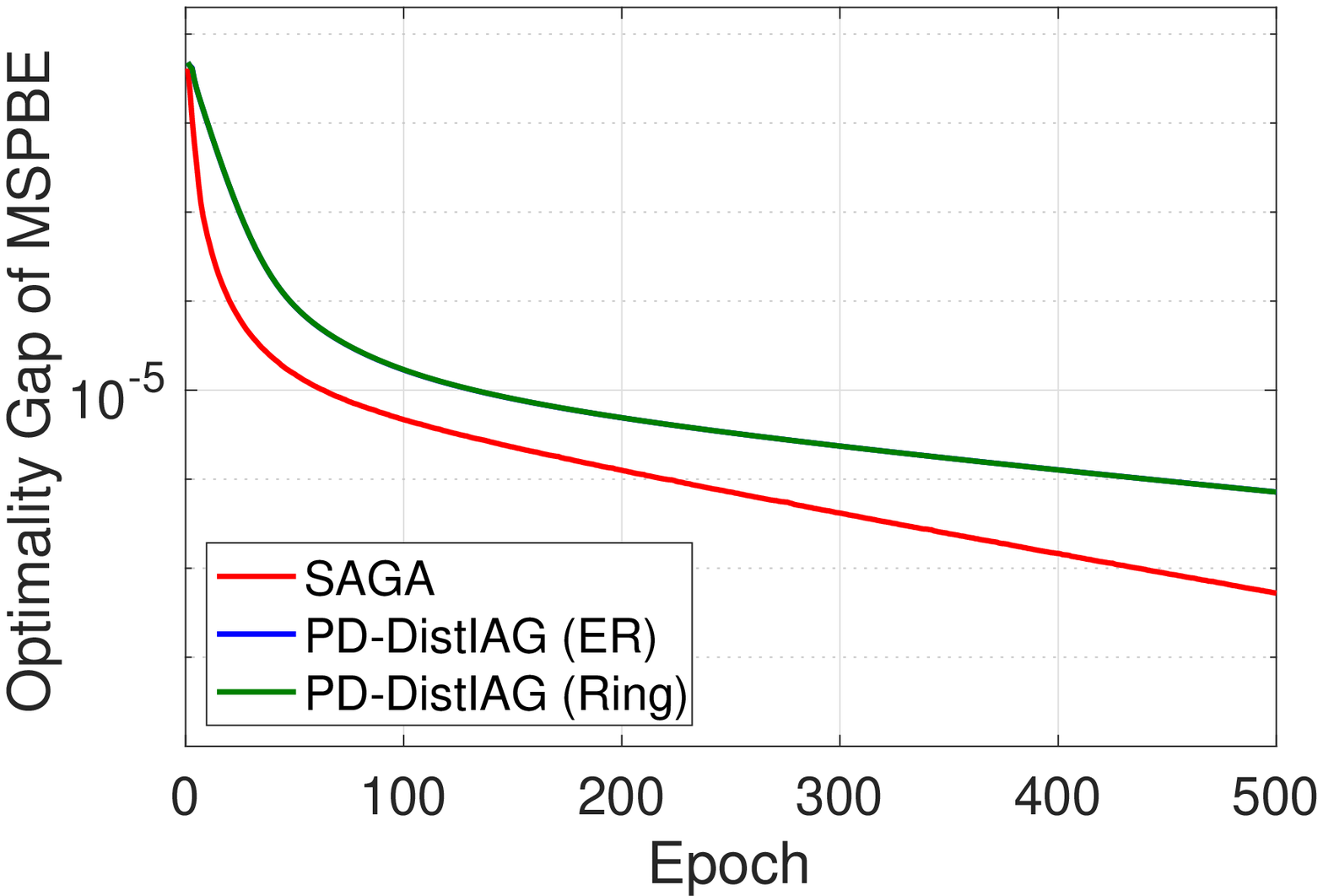}~
\includegraphics[width=.425\linewidth]{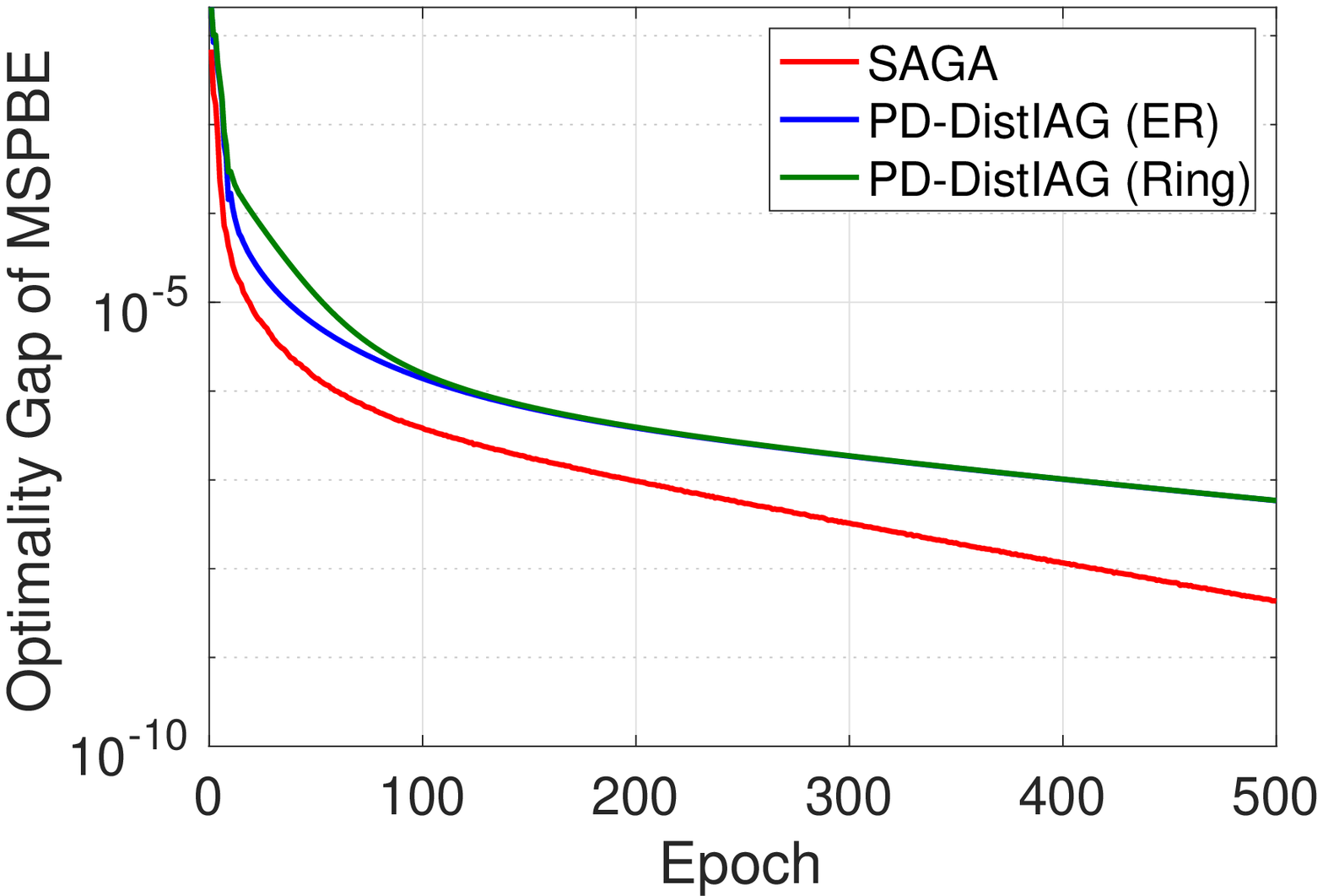}
\caption{Experiment with \texttt{mountaincar} dataset. For this problem, we 
 have  $d=300$, $M=5000$ samples, but yet there are $N=500$ agents. 
(Left) We set $\rho = 0.01$. (Right) We set $\rho = 0.1$.} \label{fig:three}
\end{figure}

Interestingly, for this example, the performances
of the ring graph and the ER graph settings are almost identical in this setting
with large sample size $M$. 
This is possible as we  
recall from Theorem~\ref{thm:main} that 
the algorithm converges at a rate of ${\cal O}(\sigma^t)$ where 
$\sigma = 1 - {\cal O}( 1 / \max\{ MN^2, M^3 \})$. As we have $M \gg N$,
the impact from the sample size $M$ becomes dominant, and is thus
insensitive to the graph's connectivity.

\bibliographystyle{abbrvnat}
\bibliography{intro.bib,to.bib,ref}

\end{document}